\documentclass{article}

\PassOptionsToPackage{numbers,sort&compress}{natbib}
\usepackage[preprint]{neurips_2022}

\usepackage[utf8]{inputenc} %
\usepackage[T1]{fontenc}    %
\usepackage{hyperref}       %
\usepackage{url}            %
\usepackage{booktabs}       %
\usepackage{amsmath}        %
\usepackage{amsfonts}       %
\usepackage{microtype}      %
\usepackage{xcolor}         %
\usepackage{graphicx}
\usepackage[capitalise,nameinlink]{cleveref}
\usepackage{paralist}
\usepackage{caption}        %
\usepackage{subcaption}     %
\usepackage{scalerel}       %
\usepackage{tikz}           %
\usepackage{bm}             %
\usepackage[export]{adjustbox} %
\usepackage{multirow}       %
\usepackage{tabularx}       %
\usepackage{wrapfig}
\usepackage{pifont}
\usepackage{physics}        %
\usepackage{rotating}

\graphicspath{{figures/}}
\usetikzlibrary{calc,backgrounds,positioning,bayesnet,graphs,arrows,shapes.geometric,fit,matrix}
\usetikzlibrary{circuits.ee.IEC} %
\crefname{section}{§}{§§}
\Crefname{section}{§}{§§}
\crefformat{equation}{(#2#1#3)}
\definecolor{red}{HTML}{E41A1C}
\definecolor{orange}{HTML}{FF7F00}
\definecolor{yellow}{HTML}{FFC020}
\definecolor{green}{HTML}{4DAF4A}
\definecolor{blue}{HTML}{377EB8}
\definecolor{purple}{HTML}{984EA3}
\newcommand{\cmark}{\textcolor{red}{\ding{51}}}
\newcommand{\xmark}{\textcolor{green}{\ding{55}}}
\newcommand{\zpres}{o} %
\newcommand{\zwhere}{l} %
\newcommand{\zwhat}{s} %
\newcommand{\ptheta}{p_\theta} %
\newcommand{\qphi}{q_\phi} %
\renewcommand{\ll}{\ell(\phi, \theta, z)} %
\renewcommand{\L}{\mathcal{L}} %
\newcommand{\sslope}{\omega_0} %
\newcommand{\gslope}{\omega_1} %
\newcommand{\xc}{x^{(c)}} %
\newcommand{\xT}{x^{(T)}} %
\newcommand{\zT}{z^{(T)}} %
\newcommand{\psic}{\psi^{(c)}} %
\newcommand{\vardbtilde}[1]{\tilde{\raisebox{0pt}[0.85\height]{$\tilde{#1}$}}} %
\newcommand{\N}{\mathcal{N}}
\newcommand{\E}{\mathbb{E}}
\setdefaultleftmargin{2ex}{2ex}{}{}{}{}
\makeatletter
\renewcommand\paragraph{\@startsection{paragraph}{4}{\z@}%
{0.5ex \@plus.1ex \@minus.1ex}%
{-1em}%
{\normalfont\normalsize\bfseries}}
\makeatother

\newif\ifnotes
\notestrue %
\ifnotes
\newcommand{\ynote}[1]{\textcolor{magenta}{\textbf{YC}: #1 }}
\newcommand{\tnote}[1]{\textcolor{blue}{\textbf{Tuan Anh}: #1 }}
\newcommand{\snote}[1]{\textcolor{yellow}{\textbf{Sid}: #1 }}
\newcommand{\jnote}[1]{\textcolor{green}{\textbf{Josh}: #1 }}
\else
\newcommand{\ynote}[1]{}
\newcommand{\tnote}[1]{}
\newcommand{\snote}[1]{}
\newcommand{\jnote}[1]{}
\fi

\title{Drawing out of Distribution with\\ Neuro-Symbolic Generative Models}

\author{%
  \hspace*{-1ex}
  Yichao Liang\(^{1,2}\),
  Joshua B. Tenenbaum\(^{2}\),
  Tuan Anh Le\footnotemark[1]\(^{\;\;,3}\)
  \& N. Siddharth\thanks{equal contribution}\(^{\;\;,4}\)\\[5pt]
  \({}^1\)University of Oxford,\,
  \({}^2\)MIT,\,
  \({}^3\)Google,\,
  \({}^4\)University of Edinburgh\\
}

\begin{document}

\maketitle
\vspace*{-\baselineskip}
\begin{abstract}
  Learning general-purpose representations from perceptual inputs is a hallmark of human intelligence.
  For example, people can write out numbers or characters, or even draw doodles, by characterizing these tasks as different instantiations of the same generic underlying process---compositional arrangements of different forms of pen strokes.
  Crucially, learning to do one task, say writing, implies reasonable competence at another, say drawing, on account of this shared process.
  We present \textbf{D}rawing \textbf{o}ut \textbf{o}f \textbf{D}istribution (DooD), a neuro-symbolic generative model of stroke-based drawing that can learn such general-purpose representations.
  In contrast to prior work, DooD operates directly on images, requires no supervision or expensive test-time inference, and performs unsupervised amortised inference with a symbolic stroke model that better enables both interpretability and generalization.
  We evaluate DooD on its ability to generalise across both data and tasks.
  We first perform zero-shot transfer from one dataset (e.g. MNIST) to another (e.g. Quickdraw), across five different datasets, and show that DooD clearly outperforms different baselines.
  An analysis of the learnt representations further highlights the benefits of adopting a symbolic stroke model.
  We then adopt a subset of the Omniglot challenge tasks, and evaluate its ability to generate new exemplars (both unconditionally and conditionally), and perform one-shot classification, showing that DooD matches the state of the art.
  Taken together, we demonstrate that DooD does indeed capture general-purpose representations across both data and task, and takes a further step towards building general and robust concept-learning systems.
\end{abstract}

\begin{wrapfigure}[13]{r}{0.35\textwidth}
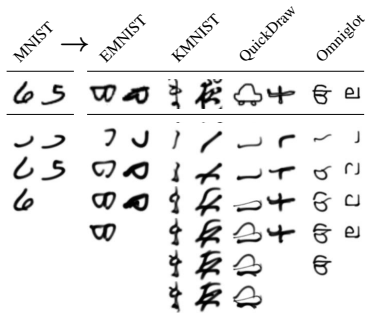

  \vspace*{-4\baselineskip}
  \centering
  \def\targetimg#1{\includegraphics[width=0.03\textwidth,trim={0 {0.6\textheight} 0 0},clip]{#1}}
\def\reconimg#1{\includegraphics[width=0.03\textwidth,trim={0 0 0 {0.097\textheight}},clip]{#1}}
\setlength\tabcolsep{0.3pt}
\begin{tabular}{@{}*{2}{c}@{\hspace*{6pt}}*{4}{*{2}{c}@{\hspace*{3pt}}}@{}}
  \multicolumn{2}{c}{\rotatebox{45}{\tiny MNIST}\,\raisebox{1ex}{\(\bm{\to}\)\!\!}}
  & \multicolumn{2}{c}{\rotatebox{45}{\tiny EMNIST}}
  & \multicolumn{2}{c}{\rotatebox{45}{\tiny KMNIST}}
  &\multicolumn{2}{c}{\rotatebox{45}{\tiny QuickDraw}}
  & \multicolumn{2}{c}{\rotatebox{45}{\tiny Omniglot}} \\[-2pt]
  \cmidrule(r){1-2} \cmidrule(r){3-10} %
  \targetimg{intro_figures/full_mn_cum_rec_MNIST_2.pdf}
  & \targetimg{intro_figures/full_mn_cum_rec_MNIST_8.pdf}
  & \targetimg{intro_figures/full_mn_cum_rec_EMNIST_2.pdf}
  & \targetimg{intro_figures/full_mn_cum_rec_EMNIST_11.pdf}
  & \targetimg{intro_figures/full_mn_cum_rec_KMNIST_4.pdf}
  & \targetimg{intro_figures/full_mn_cum_rec_KMNIST_13.pdf}
  & \targetimg{intro_figures/full_mn_cum_rec_Quickdraw_9.pdf}
  & \targetimg{intro_figures/full_mn_cum_rec_Quickdraw_27.pdf}
  & \targetimg{intro_figures/full_mn_cum_rec_Omniglot_9.pdf}
  & \targetimg{intro_figures/full_mn_cum_rec_Omniglot_30.pdf} \\[-3pt]
  \cmidrule(r){1-2} \cmidrule(r){3-10} %
  \reconimg{intro_figures/full_mn_cum_rec_nobox_MNIST_2.pdf}
  & \reconimg{intro_figures/full_mn_cum_rec_nobox_MNIST_8.pdf}
  & \reconimg{intro_figures/full_mn_cum_rec_nobox_EMNIST_2.pdf}
  & \reconimg{intro_figures/full_mn_cum_rec_nobox_EMNIST_11.pdf}
  & \reconimg{intro_figures/full_mn_cum_rec_nobox_KMNIST_4.pdf}
  & \reconimg{intro_figures/full_mn_cum_rec_nobox_KMNIST_13.pdf}
  & \reconimg{intro_figures/full_mn_cum_rec_nobox_Quickdraw_9.pdf}
  & \reconimg{intro_figures/full_mn_cum_rec_nobox_Quickdraw_27.pdf}
  & \reconimg{intro_figures/full_mn_cum_rec_nobox_Omniglot_9.pdf}
  & \reconimg{intro_figures/full_mn_cum_rec_nobox_Omniglot_30.pdf}
\end{tabular}
\setlength\tabcolsep{6pt}

  \caption{\small DooD trained on MNIST generalises to other data with no extra training. Each column denotes a target and its step-by-step reconstruction.}
  \label{fig:intro}
\end{wrapfigure}
\vspace*{-0.5\baselineskip}
\section{Introduction}
\vspace*{-0.5\baselineskip}
Humans can learn representations of data that are general-purpose and meaningful.
Being general-purpose permits effective reuse when characterizing novel observations, and being meaningful facilitates tasks like generating or classifying observations.
Key to this is a generic process for characterizing observations---inferring \emph{what} features are relevant and \emph{how} they compose to generate the observations.
For example, when observing handwritten numbers, we learn to characterise them as sequential compositions (\emph{how}) of different pen strokes (\emph{what}).
This is general-purpose as it allows characterizing novel observations, say doodles instead of numbers, simply as novel compositions of previously learnt pen strokes.
It is also meaningful since pen-strokes themselves are symbolic and interpretable.

Current computational approaches model captures important aspects of generalizability, but none of these are simultaneously efficient, reliable, interpretable, and unsupervised~\citep{lake2019omniglot}.
At one end, symbolic approaches like \citet{lake2015human} attribute generalization to an explicit hierarchical composition process involving sub-strokes, strokes, and characters, and build concomitant models that demonstrate human-like generalization abilities across different tasks.
At the other end, neural approaches like deep generative models \citep{eslami2016attend,rezende2016one,hewitt2018variational} and deep meta-learning \citep{vinyals2016matching,finn2017model,snell2017prototypical} favour scalable learning from raw perceptual data, unfettered by explicit representational biases such as strokes and their compositions.
Each comes with its own shortcomings---symbolic approaches typically need additional supervision or data processing along with expensive special-purpose inference, and neural approaches fail to generalise well and don't capture interpretable representations.
Neuro-symbolic approaches~\cite{feinman2020learning} seek to make the best of both worlds by judiciously combining neural processing of raw perceptual inputs with symbolic processing of extracted features, but typically involve a different set of trade-offs.

We present \textbf{D}rawing \textbf{o}ut \textbf{o}f \textbf{D}istribution (DooD), a neuro-symbolic generative model of stroke-based drawing that can learn general-purpose representations (\cref{fig:intro}).
Our model operates directly on images, requires no supervision, pre-processing, or expensive test-time inference, and performs efficient amortised inference with a symbolic stroke model that helps with both interpretability and generalization, setting us apart from the current state-of-the-art in neuro-symbolic approaches~\cite{feinman2020learning,hewitt2020learning}.
We evaluate on two axes
\begin{inparaenum}[(a)]
\item generalization across data, which measures how well the learnt representations can be reused to characterise out-of-distribution data, and
\item generalization across task, where we measure how useful the learnt representations are for auxiliary tasks drawn from the Omniglot challenge set~\cite{lake2015human}.
\end{inparaenum}
We show that DooD significantly outperforms baselines on generalization across datasets, highlighting the quality of the learnt representations as a factor, and on generalization across tasks, show that it outperforms neural models, while being competitive against SOTA neuro-symbolic models without requiring additional support such as supervision or data augmentation.

\vspace*{-0.5\baselineskip}
\section{Method}
\vspace*{-0.5\baselineskip}
\label{sec:method}
The framework for DooD involves a generative model over sequences of strokes and their layouts, a recognition model that conditions on a given observation to predict where to place what strokes, and an amortised variational-inference learning setup that uses these models to estimate an evidence lower bound (ELBO) as the objective.

\begin{figure}[t!]
  \centering
  \input{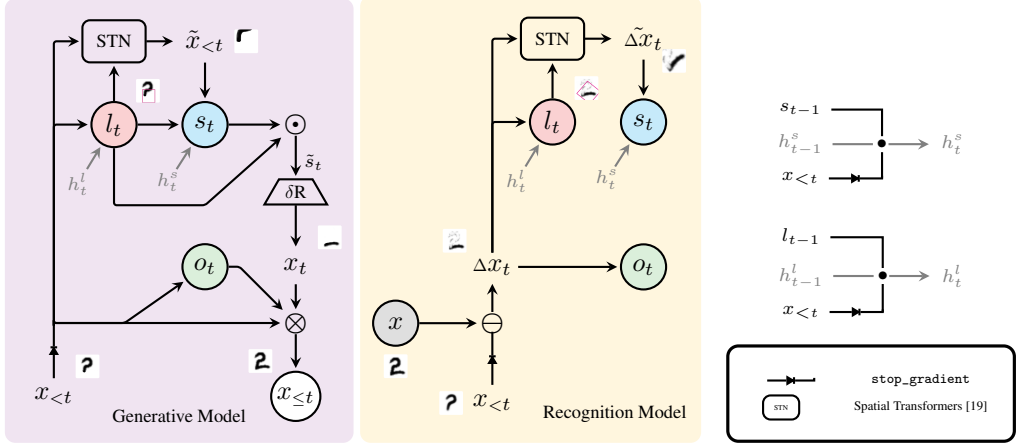}
  \caption{%
    The \emph{generative model} sequentially samples both an image location (\(l_t\)) and a corresponding stroke (\(s_t\)) at that location.
    The rendered stroke $x_t$ is composited onto intermediate rendering $x_{<t}$ to produce $x_{\leq t}$.
    A binary on/off variable ($o_t$) determines when to stop drawing.
    Differentiable rendering ($\delta R$) and differentiable affine transformations via Spatial Transformer Networks (STNs) \citep{jaderberg2015spatial} enables gradient-based learning.
    The \emph{recognition model} conditions on a residual $\Delta x_t = x - x_{<t}$ to sample where to draw next (\(l_t\)) and what to draw next (\(s_t\)), and whether to stop drawing from that point onwards ($o_t$).
    Both models are autoregressive via two (shared) RNNs with hidden states $h_t^s$ and $h_t^l$.
  }
  \label{fig:model}
\end{figure}

\vspace*{-0.5\baselineskip}
\subsection{Generative Model}
\vspace*{-0.5\baselineskip}
\label{sec:gen-model}
Conceptually, the model can be seen as drawing a figure over a sequence of steps, building up to the final image (as seen in \cref{fig:intro}).
At each step the model identifies a region of the image canvas to draw in, puts down B\'ezier curve control points within that region, renders the curve in a differentiable manner, and then composites this rendered stroke over the previously rendered canvas.
We refer to these as the \emph{layout}, \emph{stroke}, \emph{rendering}, and \emph{compositing} modules respectively (elaborated below).

The model sits on a substrate of recurrent neural networks (RNNs), one each for the layout and stroke modules, with hidden states~\(h_t^l\) and~\(h_t^s\) respectively.
The complete setup employed is depicted in \cref{fig:model} along with example values (images) for the different variables involved.

Formally, the generative model defines a joint distribution over a rendered image~\(x_{\leq T}\) following~\(T\) steps, with latent variables~\(l_t\) and~\(s_t\) that characterise a stroke's location and form, and a binary latent variable~\(o_t\) that determines how many steps are actually rendered, at each step~\(t\)
\begin{align}
  p
  &(x_{\leq T}, l_{\leq T}, s_{\leq T}, o_{\leq T}) \nonumber \\
  &= \prod_t
    p_{\text{comp}}(x_{\leq t} | x_{<t}, x_t, o_t)\;
    p_{\text{on}}(o_t | o_{<t}, x_{<t})\;
    p_{\text{stroke}}(s_t | l_t, \tilde{x}_{<t}, h_t^s)\;
    p_{\text{layout}}(l_t | x_{<t}, h_t^l).
  \label{eq:gen}
\end{align}

\paragraph{Layout Module:}
At step~\(t\), given the canvas-so-far~\(x_{<t}\) and corresponding layout-RNN hidden state~\(h_t^l\), we define the layout as a distribution over affine transforms.
This allows transforming the canvas into a ``glimpse''~\(\tilde{x}_{<t}\) using a Spatial Transformer Network (STN) \cite{jaderberg2015spatial}, which allows focussing on a particular canvas region.
The affine transform is constructed from appropriately constrained scale~(\(l_t^{\text{sc}}\)), translation~(\(l_t^{\text{tr}}\)), and rotation~(\(l_t^r\)) random variables, by employing a Gaussian Mixture Model (GMM) over the collection as
\begin{align}
  \hspace*{-1ex}
  p_{\text{layout}}(l_t | x_{<t}, h_t^l)
  &= \sum_m \alpha_m \cdot
    \N_{\text{sc},m}(l_t^{\text{sc},m} | x_{<t}, h_t^l) \cdot
    \N_{\text{tr},m}(l_t^{\text{tr},m} | x_{<t}, h_t^l) \cdot
    \N_{r,m}(l_t^{r,m} | x_{<t}, h_t^l),
    \label{eq:layout} \\
  \tilde{x}_{<t}
  &= \text{STN}(l_t, x_{<t}). \nonumber
\end{align}

\paragraph{Stroke Module:}
Given the sampled affine transform~\(l_t\), selected ``glimpse''~\(\tilde{x}_{<t}\), and corresponding stroke-RNN hidden state~\(h_t^s\), this module defines a distribution over strokes parametrised as \(D^{\text{th}}\) order B\'ezier splines, constructing a GMM for each spline control point as
\begin{align}
  p_{\text{stroke}}(s_t | l_t, \tilde{x}_{<t}, h_t^s)
  &= \prod_d \sum_k \pi_{d,k} \cdot \N_{\text{d,k}}(s_t^{d,k} | l_t, \tilde{x}_{<t}, h_t^s).
  \label{eq:stroke}
\end{align}

\paragraph{Rendering Module:}
Note that the B\'ezier spline control points sampled from the stroke module are taken to be in a canonical centered coordinate frame.
While this can help simplify learning by reducing the variation required to be captured by the stroke module, the points can't be rendered onto the canvas as is.
We situate them properly within the context of the previously determined ``glimpse'' by simply
applying the affine transform~\(l_t\) to the control points themselves as \(\tilde{s}_t = l_t \bm{\odot} s_t\).
The transformed control points now describe a stroke to be drawn over the whole canvas for
step~\(t\), which is done through a differentiable renderer~\(\delta\)R, to produce the rendered
stroke as \(x_t = \delta\text{R}(\tilde{s}_t)\).

\paragraph{Compositing Module:}
At step~\(t\) this defines a distribution over whether the model should continue drawing strokes
given the rendered canvas-so-far~\(x_{<t}\) and previous decisions~\(o_{<t}\) as
\begin{align}
  \label{eq:pres}
  p_{\text{on}}(o_t | o_{<t}, x_{<t}) = o_{t-1} \cdot \textrm{Bernoulli}(o_t | x_{<t}).
\end{align}
Once the model has decided to stop drawing, it stops permanently.
When allowed to continue, the current stroke~\(x_t\) is composited~(\(\bm{\otimes}\)) with the canvas-so-far~\(x_{<t}\) to generate the updated rendering as
\begin{align}
  \label{eq:comp}
  p_{\text{comp}}(x_{\leq t} | x_{<t}, x_t, o_t)
  = \textrm{Laplace}(x_{\leq t} | (x_{<t} \bm{\otimes} x_t), o_t = 1).
\end{align}

\subsection{Recognition Model}
\label{sec:rec-model}

As with prior approaches, we construct an approximate posterior to facilitate learning with amortised variational inference.
Using the same notation from the generative model section, we define
\begin{align}
  q(l_{< T}, s_{< T}, o_{\leq T} | x)
  &= q_{\text{on}}(o_{T}|\Delta x_{T})
  \prod_t
    q_{\text{layout}}(l_t | \Delta x_t, h_t^l) \cdot
    q_{\text{stroke}}(s_t | \tilde{\Delta x_t}, h_t^s) \cdot
    q_{\text{on}}(o_t | \Delta x_t).
  \label{eq:rec}
\end{align}
There are a couple of things worth noting.
First, where the generative model made heavy use of the canvas-so-far~\(x_{<t}\), the recognition model primarily uses the \emph{residual}~\(\Delta x_t = x - x_{<t}\).
Second, being given the target observation~\(x\) itself, the information available to the layout and stroke modules is quite different.
In the generative model, these modules have to speculate where and what stroke to draw, but in the recognition model, their task is simply to isolate a part of the drawing (``glimpse'') and fit a spline to that.

As a consequence of these different characteristics, the distributions over layout and strokes in the recognition model do not need to be as flexible as the generative model---locating a curve in the residual and fitting it with a spline does not typically involve much ambiguity.
To factor this in, and have the variational objective be reasonable, we define corresponding
distributions in the recognition model using just a single component of the corresponding GMMs in the generative model as
\begin{align}
  q_{\text{layout}}(l_t | \Delta x_{t}, h_t^l)
  &= \N_{\text{sc}}(l_t^{\text{sc}} | \Delta x_{t}, h_t^l) \cdot
    \N_{\text{tr}}(l_t^{\text{tr}} | \Delta x_{t}, h_t^l) \cdot
    \N_r(l_t^r | \Delta x_{<t}, h_t^l),
    \label{eq:layout-r} \\
  \tilde{\Delta x}_{t}
  &= \text{STN}(l_t, \Delta x_{t}), \nonumber \\
  q_{\text{stroke}}(s_t | \tilde{\Delta x}_{t}, h_t^s)
  &= \prod_d \N_{\text{d}}(s_t^{d} | \tilde{\Delta x}_{t}, h_t^s).
\end{align}

\vspace*{-\baselineskip}
\subsection{Learning}
\label{sec:learning}

Having defined the generative and recognition models, we now bring them together in order to construct the variational objective that will enable learning both models simultaneously from data.
\begin{align}
  \log p(x)
  &\geq \E_{q(l_{\leq T}, s_{\leq T}, o_{\leq T+1} | x)} \left[
    \log \frac{p(x_{\leq T}, l_{\leq T}, s_{\leq T}, o_{\leq T+1})}{q(l_{\leq T}, s_{\leq T}, o_{\leq T+1} | x)}
    \right]
  \label{eq:elbo}
\end{align}
Note that except for the stopping criterion~\(o_t\) which is a Bernoulli random variable, all other distributions employed are reparametrizable.
In order to construct an effective variational objective with this discrete variable, we employ a control variate method, NVIL~\cite{mnih2014neural}, that helps reduce the variance of the standard REINFORCE estimator, as is also done in related work~\cite{eslami2016attend}.

Furthermore, in order to ensure that the ELBO objective is appropriately balanced, we employ additional weighting~\(\beta\) for the KL-divergence over stopping criterion~\(o_t\) within the objective~\cite{bowman2016generating,higgins2016beta}.
This weight plays a crucial role as a mismatch could result in the model either stopping too early or too late, resulting in incomplete or incorrect figures respectively.

\vspace*{-0.3\baselineskip}
\section{Experiments}
\label{sec:experiments}

We wish to understand how well DooD generalises across both datasets (\cref{sec:across_dataset_generalization}) and tasks (\cref{sec:across_task_generalization}).
For across-dataset generalization, we train DooD and Attend-Infer-Repeat (AIR)~\citep{eslami2016attend}\footnote{specifically Difference-AIR, which uses guided execution and performs much better than vanilla AIR}, an unsupervised part-based model, on each of five stroke-based image datasets
\begin{inparaenum}[(i)]
\item MNIST (handwritten digits) \citep{lecun1998mnist},
\item EMNIST (handwritten digits and letters) \citep{cohen2017emnist},
\item KMNIST (cursive Japanese characters) \citep{clanuwat2018deep},
\item Quickdraw (doodles) \citep{ha2017neural}, and
\item Omniglot (handwritten characters from multiple alphabets) \citep{lake2015human},
\end{inparaenum}
and evaluate how well the model generalises to unseen exemplars both within the same dataset and across other datasets.
We find that DooD significantly outperforms AIR, which from ablation studies, is attributed to explicit stroke modelling and guided execution.
Note that we only compare against a fully-unsupervised approach since most datasets do not provide additional data in the form of stroke labels (as required elsewhere~\cite{feinman2020learning}).
For across-task generalization, we primarily focus on Omniglot and evaluate on three out of the five challenge tasks for this dataset~\citep{lake2019omniglot}, which include contextual generation and classification.
We find that our model outperforms unsupervised baselines where appropriate, and is competitive against SOTA neuro-symbolic models without requiring additional support in the form of supervision or data augmentation.
We include exact details about datasets (\cref{app:dataset}), our model and the baselines (\cref{app:model}), the training procedure (\cref{app:training}), and the evaluation procedure (\cref{app:evaluation}) in the supplementary material.

\vspace*{-0.3\baselineskip}
\subsection{Across-Dataset Generalization}
\vspace*{-0.3\baselineskip}
\label{sec:across_dataset_generalization}

\begin{figure}[t!]
  \scalebox{0.87}{\def\targetimg#1{\includegraphics[width=0.03\textwidth,trim={0 {0.6\textheight} 0 0},clip]{#1}}
\def\reconimg#1{\includegraphics[width=0.03\textwidth,trim={0 0 0 {0.097\textheight}},clip]{#1}}
\setlength\tabcolsep{1pt}
\begin{tabular}{@{}c@{\quad}*{4}{c}@{\hspace{20pt}}*{4}{*{4}{c}@{\hspace{10pt}}}@{\hspace{-10pt}}}
  (a)
  & \multicolumn{4}{c}{\quad MNIST \quad\(\bm{\to}\)\,}
  & \multicolumn{4}{c}{EMNIST}
  & \multicolumn{4}{c}{KMNIST}
  & \multicolumn{4}{c}{QuickDraw}
  & \multicolumn{4}{c}{Omniglot} \\
  \midrule
  $x$
  & \targetimg{mnist_trans_recon/full_cum_MNIST_0.pdf}
  & \targetimg{mnist_trans_recon/full_cum_MNIST_2.pdf}
  & \targetimg{mnist_trans_recon/full_cum_MNIST_23.pdf}
  & \targetimg{mnist_trans_recon/full_cum_MNIST_25.pdf}
  & \targetimg{mnist_trans_recon/full_cum_EMNIST_6.pdf}
  & \targetimg{mnist_trans_recon/full_cum_EMNIST_11.pdf}
  & \targetimg{mnist_trans_recon/full_cum_EMNIST_12.pdf}
  & \targetimg{mnist_trans_recon/full_cum_EMNIST_13.pdf}
  & \targetimg{mnist_trans_recon/full_cum_KMNIST_2.pdf}
  & \targetimg{mnist_trans_recon/full_cum_KMNIST_10.pdf}
  & \targetimg{mnist_trans_recon/full_cum_KMNIST_11.pdf}
  & \targetimg{mnist_trans_recon/full_cum_KMNIST_13.pdf}
  & \targetimg{mnist_trans_recon/full_cum_Quickdraw_14.pdf}
  & \targetimg{mnist_trans_recon/full_cum_Quickdraw_16.pdf}
  & \targetimg{mnist_trans_recon/full_cum_Quickdraw_26.pdf}
  & \targetimg{mnist_trans_recon/full_cum_Quickdraw_30.pdf}
  & \targetimg{mnist_trans_recon/full_cum_Omniglot_0.pdf}
  & \targetimg{mnist_trans_recon/full_cum_Omniglot_3.pdf}
  & \targetimg{mnist_trans_recon/full_cum_Omniglot_11.pdf}
  & \targetimg{mnist_trans_recon/full_cum_Omniglot_19.pdf} \\
  \midrule
  \raisebox{8ex}{\rotatebox[origin=c]{90}{DooD}}
  & \reconimg{mnist_trans_recon/full_cum_MNIST_0.pdf}
  & \reconimg{mnist_trans_recon/full_cum_MNIST_2.pdf}
  & \reconimg{mnist_trans_recon/full_cum_MNIST_23.pdf}
  & \reconimg{mnist_trans_recon/full_cum_MNIST_25.pdf}
  & \reconimg{mnist_trans_recon/full_cum_EMNIST_6.pdf}
  & \reconimg{mnist_trans_recon/full_cum_EMNIST_11.pdf}
  & \reconimg{mnist_trans_recon/full_cum_EMNIST_12.pdf}
  & \reconimg{mnist_trans_recon/full_cum_EMNIST_13.pdf}
  & \reconimg{mnist_trans_recon/full_cum_KMNIST_2.pdf}
  & \reconimg{mnist_trans_recon/full_cum_KMNIST_10.pdf}
  & \reconimg{mnist_trans_recon/full_cum_KMNIST_11.pdf}
  & \reconimg{mnist_trans_recon/full_cum_KMNIST_13.pdf}
  & \reconimg{mnist_trans_recon/full_cum_Quickdraw_14.pdf}
  & \reconimg{mnist_trans_recon/full_cum_Quickdraw_16.pdf}
  & \reconimg{mnist_trans_recon/full_cum_Quickdraw_26.pdf}
  & \reconimg{mnist_trans_recon/full_cum_Quickdraw_30.pdf}
  & \reconimg{mnist_trans_recon/full_cum_Omniglot_0.pdf}
  & \reconimg{mnist_trans_recon/full_cum_Omniglot_3.pdf}
  & \reconimg{mnist_trans_recon/full_cum_Omniglot_11.pdf}
  & \reconimg{mnist_trans_recon/full_cum_Omniglot_19.pdf} \\
  \midrule
  \raisebox{8ex}{\rotatebox[origin=c]{90}{AIR}}
  & \reconimg{mnist_trans_recon/dair_cum_MNIST_0.pdf}
  & \reconimg{mnist_trans_recon/dair_cum_MNIST_2.pdf}
  & \reconimg{mnist_trans_recon/dair_cum_MNIST_23.pdf}
  & \reconimg{mnist_trans_recon/dair_cum_MNIST_25.pdf}
  & \reconimg{mnist_trans_recon/dair_cum_EMNIST_6.pdf}
  & \reconimg{mnist_trans_recon/dair_cum_EMNIST_11.pdf}
  & \reconimg{mnist_trans_recon/dair_cum_EMNIST_12.pdf}
  & \reconimg{mnist_trans_recon/dair_cum_EMNIST_13.pdf}
  & \reconimg{mnist_trans_recon/dair_cum_KMNIST_2.pdf}
  & \reconimg{mnist_trans_recon/dair_cum_KMNIST_10.pdf}
  & \reconimg{mnist_trans_recon/dair_cum_KMNIST_11.pdf}
  & \reconimg{mnist_trans_recon/dair_cum_KMNIST_13.pdf}
  & \reconimg{mnist_trans_recon/dair_cum_Quickdraw_14.pdf}
  & \reconimg{mnist_trans_recon/dair_cum_Quickdraw_16.pdf}
  & \reconimg{mnist_trans_recon/dair_cum_Quickdraw_26.pdf}
  & \reconimg{mnist_trans_recon/dair_cum_Quickdraw_30.pdf}
  & \reconimg{mnist_trans_recon/dair_cum_Omniglot_0.pdf}
  & \reconimg{mnist_trans_recon/dair_cum_Omniglot_3.pdf}
  & \reconimg{mnist_trans_recon/dair_cum_Omniglot_11.pdf}
  & \reconimg{mnist_trans_recon/dair_cum_Omniglot_19.pdf} \\
  \midrule
  \raisebox{8ex}{\rotatebox[origin=c]{90}{DooD-\(\delta\)R}}
  & \reconimg{mnist_trans_recon/ma2_cum_MNIST_0.pdf}
  & \reconimg{mnist_trans_recon/ma2_cum_MNIST_2.pdf}
  & \reconimg{mnist_trans_recon/ma2_cum_MNIST_23.pdf}
  & \reconimg{mnist_trans_recon/ma2_cum_MNIST_25.pdf}
  & \reconimg{mnist_trans_recon/ma2_cum_EMNIST_6.pdf}
  & \reconimg{mnist_trans_recon/ma2_cum_EMNIST_11.pdf}
  & \reconimg{mnist_trans_recon/ma2_cum_EMNIST_12.pdf}
  & \reconimg{mnist_trans_recon/ma2_cum_EMNIST_13.pdf}
  & \reconimg{mnist_trans_recon/ma2_cum_KMNIST_2.pdf}
  & \reconimg{mnist_trans_recon/ma2_cum_KMNIST_10.pdf}
  & \reconimg{mnist_trans_recon/ma2_cum_KMNIST_11.pdf}
  & \reconimg{mnist_trans_recon/ma2_cum_KMNIST_13.pdf}
  & \reconimg{mnist_trans_recon/ma2_cum_Quickdraw_14.pdf}
  & \reconimg{mnist_trans_recon/ma2_cum_Quickdraw_16.pdf}
  & \reconimg{mnist_trans_recon/ma2_cum_Quickdraw_26.pdf}
  & \reconimg{mnist_trans_recon/ma2_cum_Quickdraw_30.pdf}
  & \reconimg{mnist_trans_recon/ma2_cum_Omniglot_0.pdf}
  & \reconimg{mnist_trans_recon/ma2_cum_Omniglot_3.pdf}
  & \reconimg{mnist_trans_recon/ma2_cum_Omniglot_11.pdf}
  & \reconimg{mnist_trans_recon/ma2_cum_Omniglot_19.pdf} \\
  \midrule
  \raisebox{8ex}{\rotatebox[origin=c]{90}{DooD-EG}}
  & \reconimg{mnist_trans_recon/ma3_cum_MNIST_0.pdf}
  & \reconimg{mnist_trans_recon/ma3_cum_MNIST_2.pdf}
  & \reconimg{mnist_trans_recon/ma3_cum_MNIST_23.pdf}
  & \reconimg{mnist_trans_recon/ma3_cum_MNIST_25.pdf}
  & \reconimg{mnist_trans_recon/ma3_cum_EMNIST_6.pdf}
  & \reconimg{mnist_trans_recon/ma3_cum_EMNIST_11.pdf}
  & \reconimg{mnist_trans_recon/ma3_cum_EMNIST_12.pdf}
  & \reconimg{mnist_trans_recon/ma3_cum_EMNIST_13.pdf}
  & \reconimg{mnist_trans_recon/ma3_cum_KMNIST_2.pdf}
  & \reconimg{mnist_trans_recon/ma3_cum_KMNIST_10.pdf}
  & \reconimg{mnist_trans_recon/ma3_cum_KMNIST_11.pdf}
  & \reconimg{mnist_trans_recon/ma3_cum_KMNIST_13.pdf}
  & \reconimg{mnist_trans_recon/ma3_cum_Quickdraw_14.pdf}
  & \reconimg{mnist_trans_recon/ma3_cum_Quickdraw_16.pdf}
  & \reconimg{mnist_trans_recon/ma3_cum_Quickdraw_26.pdf}
  & \reconimg{mnist_trans_recon/ma3_cum_Quickdraw_30.pdf}
  & \reconimg{mnist_trans_recon/ma3_cum_Omniglot_0.pdf}
  & \reconimg{mnist_trans_recon/ma3_cum_Omniglot_3.pdf}
  & \reconimg{mnist_trans_recon/ma3_cum_Omniglot_11.pdf}
  & \reconimg{mnist_trans_recon/ma3_cum_Omniglot_19.pdf}
\end{tabular}

}
  \hspace{2pt}
  \begin{tabular}{@{}c@{}}
    \includegraphics[width=0.22\linewidth]{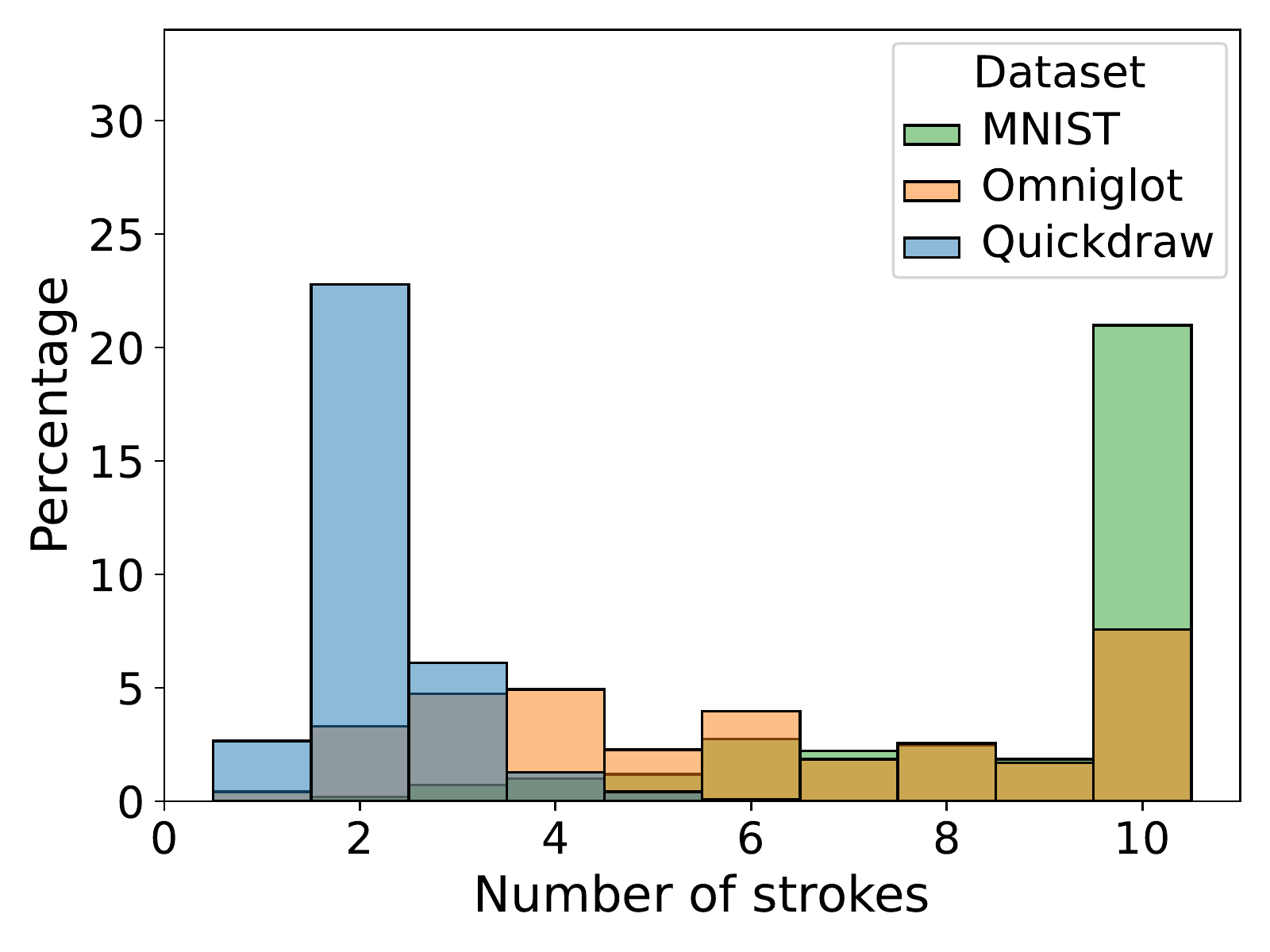} \\[-1ex]
    \footnotesize (b) DooD \\[5ex]
    \includegraphics[width=0.22\linewidth]{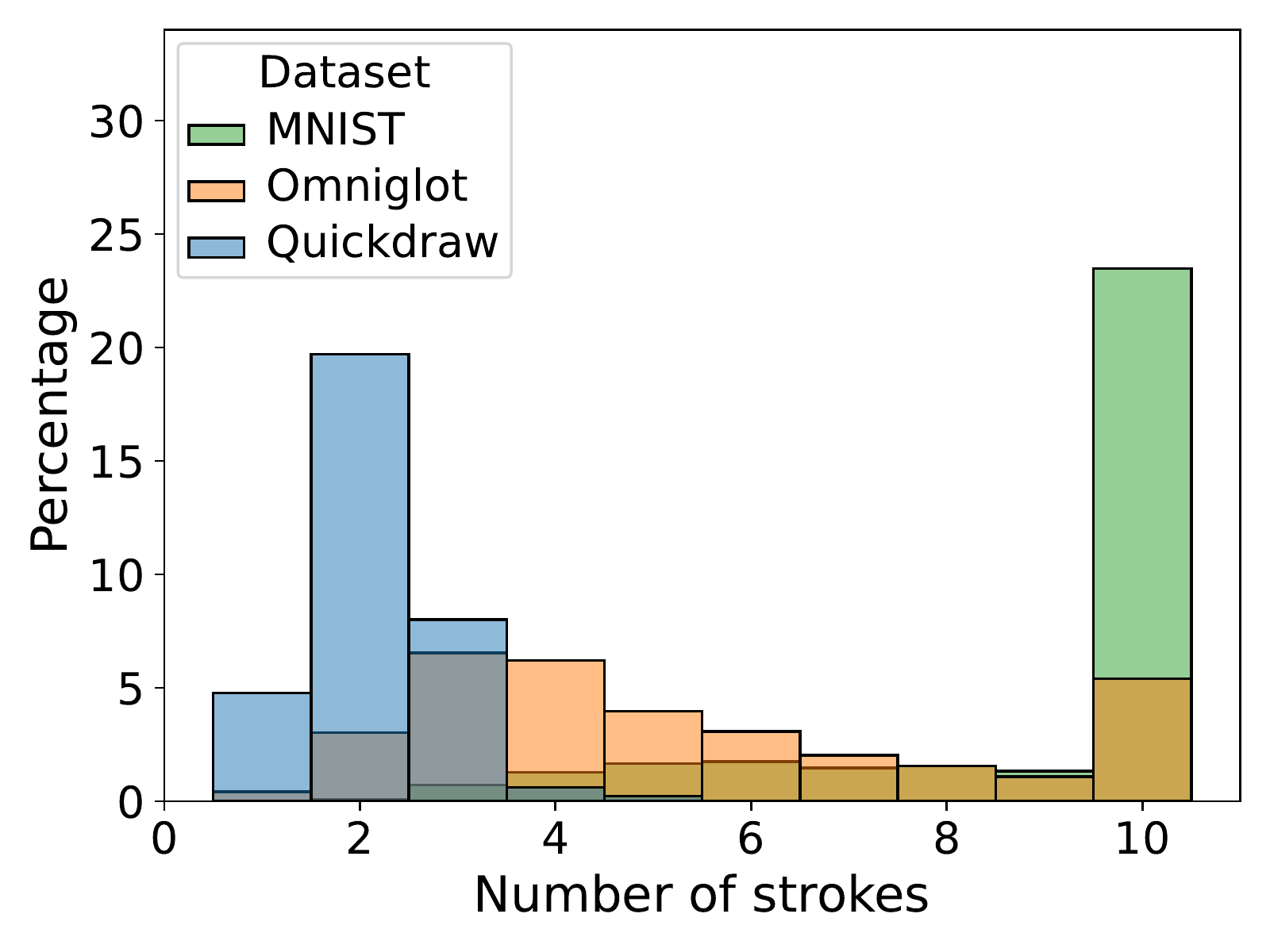} \\[-1ex]
    \footnotesize (c) Difference AIR \cite{eslami2016attend} \\[5ex]
    \includegraphics[width=0.22\linewidth]{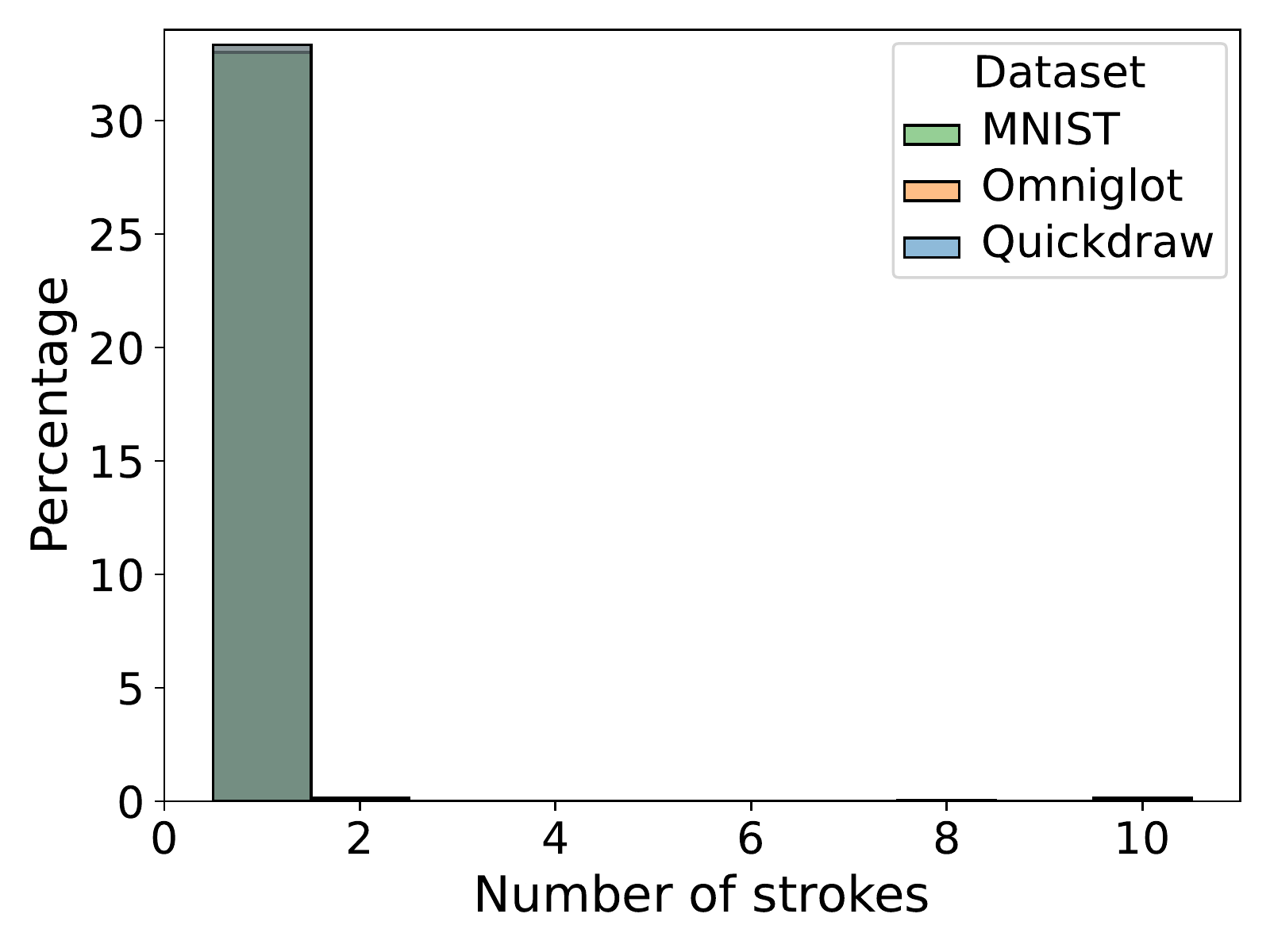}  \\[-1ex]
    \footnotesize (d) Vanilla AIR \cite{eslami2016attend}
  \end{tabular}
  \caption{(a) Our model generalises better than AIR.
    Our model trained on MNIST reconstructs characters from all other four datasets while the baseline AIR model's reconstructions are often inaccurate, blurry or incomplete.
    Explicit stroke parametrization ($\delta$R) and execution-guided inference (EG) are responsible for this generalization which degrades when using our model without either of these components.
    (b-d) Both DooD and Difference AIR (AIR elsewhere) trained on MNIST generalise to using more strokes unlike Vanilla AIR which doesn't have execution-guided inference.}
  \label{fig:mnist_generalization}
\end{figure}

\paragraph{MNIST-trained transfer.}
To understand how our model and AIR generalise to new datasets, we look at sequential reconstructions (\cref{fig:mnist_generalization}).
We train on MNIST and show sample reconstructions from all five datasets without fine tuning.
Each model renders one step at a time by rendering latent parses of increasing length, allowing us to evaluate and compare the performance of part decomposition and inference.
Note that we limit the maximum number of strokes to 6 throughout all experiments.

Our model reconstructs in-distribution images perfectly and out-of-distribution images near-perfectly while using fewer strokes for simpler datasets (e.g. MNIST) and more strokes for more complex datasets (e.g. Omniglot).
While the AIR baseline also uses an appropriate number of steps for more complex datasets, the reconstructions degrade significantly for out-of-distribution images---they are blurry (e.g. the car \& motorbike in QuickDraw), strokes go missing (e.g. the second KMNIST image) or the reconstructions are inaccurate (e.g. the last Omniglot character).

\textbf{Ablation studies.}
To better understand why our model generalises well, we evaluate two further variants of DooD that ablate a key component each: an explicit spline decoder (DooD-\(\delta\)R) and execution-guided inference (DooD-EG) (\cref{fig:mnist_generalization}a).
In the model without an explicit spline decoder, we replace the differentiable spline renderer by a neural network decoder similar to AIR.
This model still differs from the AIR in terms of the learnable sequential prior and the fact that we enforce explicit constrains over the latent variable ranges---e.g. enforcing the mean of the control-point Gaussian to not stray too far away from the image frame.
In the variant without guided execution, we do not perform intermediate rendering, removing the direct dependence of the generative model and the recognition model on the canvas-so-far $x_{<t}$ and the residual $\Delta x_t$.

Both the explicit spline decoder and the guided execution prove to be important.
Without the explicit spline decoder (DooD-\(\delta\)R), the reconstruction quality suffers---the strokes are blurry (e.g. first three QuickDraw images), strokes go missing (e.g. the last EMNIST image), or the reconstructions are inaccurate (e.g. the last Omniglot character is interpreted as a ``9'' due to overfitting).
However, even without the explicit spline decoder, the model learns to be parsimonious, using fewer strokes to reconstruct simpler images (\cref{fig:mnist_generalization}b-d).
On the other hand, without guided execution (DooD-EG), the model is unable to be selective with the number of strokes, always using the maximum allowed number.
And while the reconstructions are better than AIR and DooD-\(\delta\)R, it still shows instances of missing strokes (e.g. some Omniglot characters).
Note that although we use the canvas-so-far in a manner that disallows gradients (\texttt{stop\textunderscore{}gradient}), just providing it as a conditioning variable for the different components (layout, stroke, RNN hidden states) has a tangible effect.

\begin{figure}[!t]
  \centering
  \begin{tikzpicture}
    \node[anchor=south west,inner sep=0] (img) at (0, 0) {%
      \includegraphics[width=1.0\textwidth]{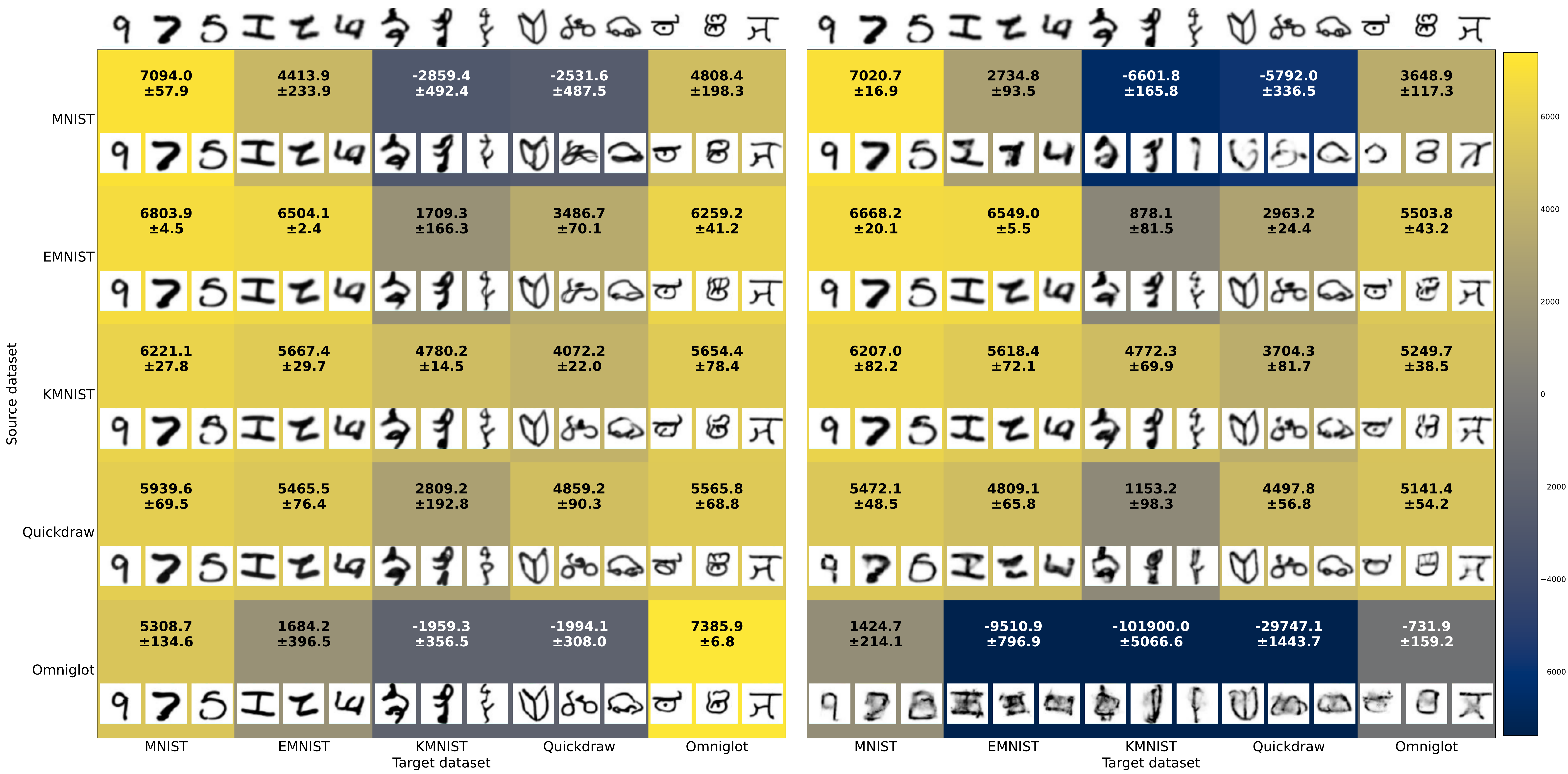}
    };
    \begin{scope}[x={(img.south east)},y={(img.north west)}]
      \draw[purple, ultra thick, rounded corners=1pt] (0.513, 0.049) rectangle (0.955, 0.225);
    \end{scope}
  \end{tikzpicture}
  \vspace*{0.7ex}
  \caption{%
    When training on a ``source'' dataset and testing on another ``target'' dataset, our model, DooD, (left) has a higher log marginal likelihood (values in each cell) than AIR (right).
    Given targets on top of the tables, DooD's reconstructions (images in each cell) are high quality when transferring out of distribution, unlike AIR which often struggles.
    Training on MNIST or Omniglot as a source dataset leads to worse transfer (the corresponding rows are the darkest) due to a larger distribution shift.
    Particularly AIR fails when trained on Omniglot using a Laplace likelihood (standard across all other model-dataset combination for good reconstructions), due to which we employ a Gaussian likelihood just for Omniglot-trained AIR (highlighted in purple).}
  \label{fig:cross_ds_mll_eval}
\end{figure}

\paragraph{Quantifying zero-shot transfer.}
We look at how DooD and AIR trained on each of the five datasets transfers to each other dataset to further understand how our model generalises.
Models are trained on each ``source'' dataset and tested on each ``target'' dataset, resulting in a $5 \times 5$ table for each model (\cref{fig:cross_ds_mll_eval}).
Each cell shows the log marginal likelihood of the target dataset using the model trained on the source dataset, estimated using the importance weighted autoencoder (IWAE) objective \citep{burda2015importance} with 200 samples (mean and standard deviation over five runs).
We also show reconstructions obtained by running the model trained on the source dataset on a few examples from the target dataset.

Our model generalises significantly better than AIR across datasets (off diagonal cells), while also performing better within dataset (diagonal cells).
For both models, the values on the diagonal are the highest in any given column, suggesting that not training on directly on the target dataset results in a worse performance, as expected.
For both models, the row values for MNIST and Omniglot are lower than in other rows, indicating that transfer learning performance is the worst when the source dataset is MNIST or Omniglot---potentially due to a larger distribution shift since MNIST has low diversity and Omniglot has little to no variation in stroke thickness, in contrast to the other datasets.
However, we note that our reconstructions are high quality despite transferring out of distribution, unlike reconstructions from AIR which are qualitatively worse.
For example, when transferring from simple datasets (MNIST), AIR makes incomplete, incorrect and blurry reconstructions, as we have seen before, while AIR trained on complex datasets like Omniglot results in blurry reconstructions for both in-distribution and out-of-distribution datasets.
Furthermore, AIR fails when trained on Omniglot using a Laplace likelihood (used as standard across all other model-dataset combination).
We thus employ a Gaussian likelihood just for Omniglot-trained AIR, and highlight it as an outlier.

\begin{figure}[h!]
  \centering
  \includegraphics[width=0.49\linewidth]{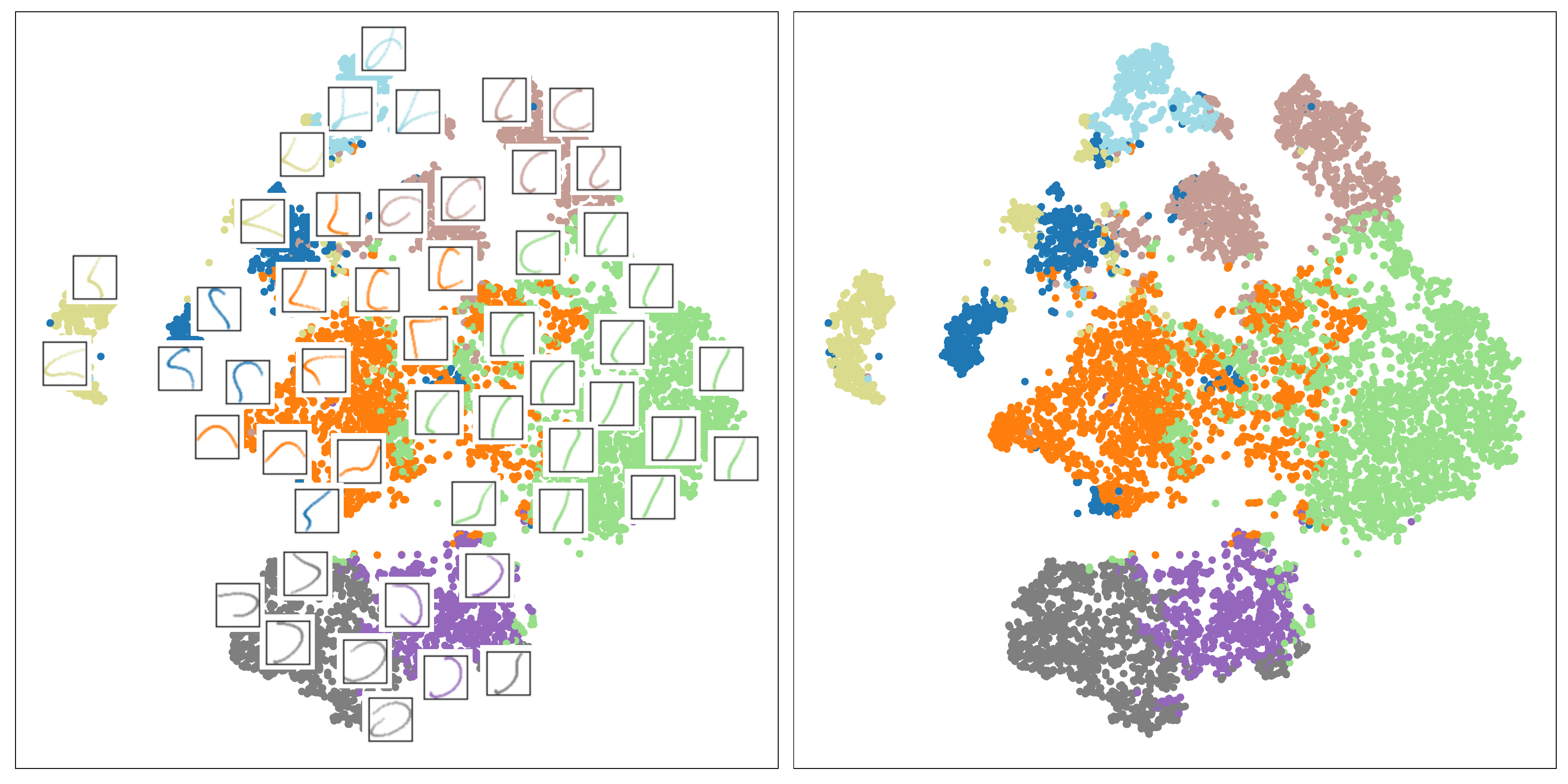}
  \includegraphics[width=0.49\linewidth]{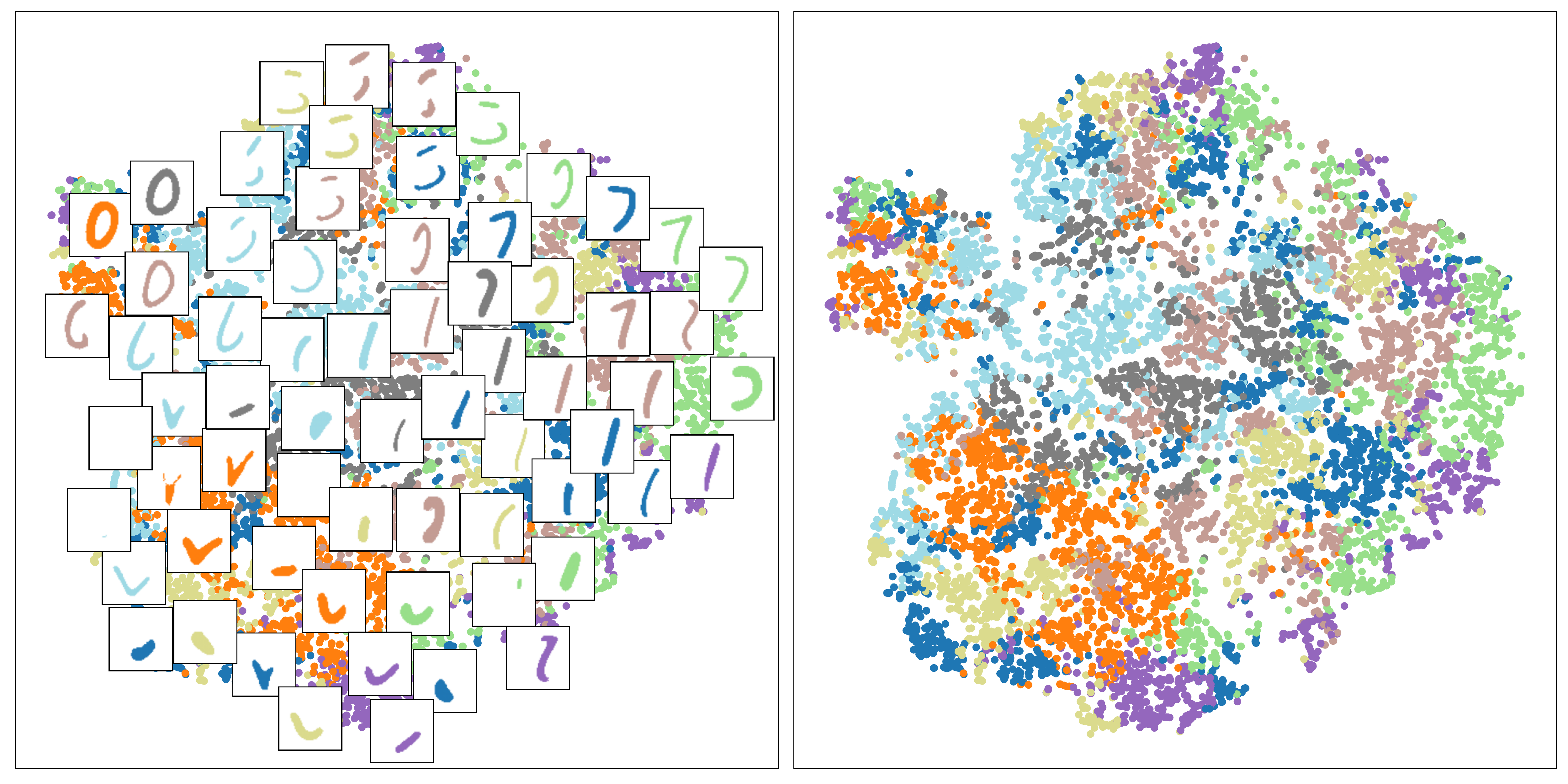}
\caption{Clusters of inferred strokes for DooD (left) and inferred part representations for AIR (right) overlaid on a t-SNE plot. DooD's representation clusters to more semantically meaningful parts as indicated by better formed clusters.}
\label{fig:lv_clf_vis}
\end{figure}

\paragraph{Understanding learned representations.}
To better understand DooD's generalization ability, we investigate its learnt representations by clustering the inferred strokes using $k$-means clustering ($k = 8$), and study the clusters both qualitatively and quantitatively.
For AIR, we cluster the corresponding part-representation latents.
We then visualise things, using a t-SNE plot (\cref{fig:lv_clf_vis}) of the clusters, with exemplar strokes overlaid.
We find that DooD has better-clustered representations, with clusters denoting largely distinct types of strokes---e.g., clusters for a ``/'', ``c'', and its horizontally flipped version.
In contrast, the clusters from AIR are less sensible with some clusters even capturing full characters (``0''), comprising multiple strokes.
There are also clusters which contain visually different strokes, and many visually similar strokes are assigned to different clusters.
Quantitatively, following \citet{aksan2020cose}, we found DooD's better cluster consistency is reflected in a higher Silhouette Coefficient \citep{rousseeuw1987silhouettes} than AIR (0.21 for DooD, 0.11 for AIR).

\subsection{Across-Task Generalization}
\label{sec:across_task_generalization}

Here, we focus on a subset of the Omniglot challenge tasks \citep{lake2015human}, to evaluate our model's utility on a range of auxiliary tasks that it was not trained to do.
Despite much progress in deep generative modelling, relevant models are still not fully task general and often result in unrealistic (e.g. blurry) samples \citep{lake2019omniglot,feinman2020learning}.
DooD combines handling raw perceptual inputs with the compositional structure of strokes,
which we evaluate on three out of the five Omniglot challenge tasks: unconditional generation, conditional generation, and one-shot classification.
We compare against AIR and a state-of-the-art neuro-symbolic model (GNS \citep{feinman2020learning}), where relevant.
Note that GNS requires stroke and character-class supervision and practically, at least for now, only applies to Omniglot.

\paragraph{Unconditional generation.}
DooD generates realistic unconditional samples of all datasets (\cref{fig:uncon_gen}), indicating that the model has learned the high-level characteristics of each dataset.
The strokes are sharp, and the stroke structure composes into realistic images from each dataset.
For example, there are clear digits in the MNIST samples, there are recognizable objects (cars, bicycles, glasses, and smileys) in the QuickDraw samples, and the samples for EMNIST, KMNIST and Omniglot can be easily recognised as possible instances coming from those datasets.
It generates samples of comparable fidelity to GNS without requiring any supervision, and as evaluated using the Fréchet inception distance (FID) \citep{heusel2017gans} (smaller is better), outperforms GNS (0.051 versus 0.133).

The key to being able to generate realistic prior samples is the learnable sequential prior and the symbolic latent representation.
AIR doesn't have a sequential prior, so although it is possible to get good reconstructions, it is impossible for it to generate realistic unconditional samples.

\begin{figure}[!b]
  \vspace*{-1\baselineskip}
  \def\gencropimg#1{\includegraphics[width=14ex, trim={0 0 {2.2\textwidth} 0},clip]{#1}}
  \centering\scriptsize
  \begin{tabular}{@{}*{7}{c@{\hspace*{10pt}}}@{\hspace*{-10pt}}}
    MNIST
    & EMNIST
    & KMNIST
    & QuickDraw
    & Omniglot (DooD)
    & Omniglot (GNS\cite{feinman2020learning})
    & Omniglot (True)\\
    $.134\pm.013$
    & $.137\pm.006$
    & $.123\pm.020$
    & $.084\pm.009$
    & $.051\pm.007$
    & $.133\pm.007$
    & $.025\pm.004$\\
    \midrule
    \gencropimg{uncon_generation/uncon_gen_mn.pdf}
    & \gencropimg{uncon_generation/uncon_gen_em.pdf}
    & \gencropimg{uncon_generation/uncon_gen_km.pdf}
    & \gencropimg{uncon_generation/uncon_gen_qd.pdf}
    & \gencropimg{uncon_generation/uncon_gen_om.pdf}
    & \gencropimg{uncon_generation/uncon_gen_gns.pdf}
    & \gencropimg{uncon_generation/omniglot_sample.pdf}
  \end{tabular}
  \caption{DooD generates high quality unconditional character samples for all datasets which are visually indistinguishable from the real characters as it successfully captures the layout of strokes and their forms. 
  Omniglot samples are compared to GNS \citep{feinman2020learning} and real samples.
  Numbers denote Fréchet inception distance (FID), with smaller being better (mean $\pm$ 1 std. over 5 runs).}
  \label{fig:uncon_gen}
\end{figure}

\paragraph{Character-conditioned generation.}

In order to generate new exemplars of the same Omniglot character, we follow \citet{lake2015human,feinman2020learning} and extend our model to a hierarchical generative model of an abstract character ``type'' or ``template'' that generates a concrete instance of a character ``token'', which is rendered out to an image. 
We consider the previously used latent variables as the type latent variable and introduce a token model which conditions on the type latent variable.
The token model introduces (i) a drawing noise represented by adding a Gaussian noise with fixed standard deviations to spline control points and (ii) and an affine transformation on the noise perturbed points, whose parameters are also sampled from a Gaussian distribution (described in \cref{app:token_model}).
To sample a new exemplar of a character, we first sample the type variable from our recognition model, and produce different exemplars by sampling and rendering different token variables given this type variable.
Distinctly from \citep{lake2015human,feinman2020learning}, the parameters of the token model are not learned through supervision but simply set to a sensible value by examining the noise level in the output -- any learned statistics can be straightforwardly plugged in.

\begin{figure}[!t]
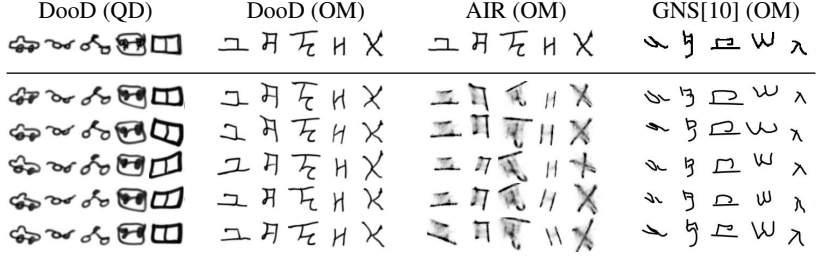

  \def\concropgt#1{\includegraphics[width=15ex, trim={0 {1\textheight} {4.5\textwidth} 0},clip]{#1}}
  \def\concropimg#1{\includegraphics[width=15ex, trim={0 0 {4.5\textwidth} {.6\textheight}},clip]{#1}}
  \centering
  \begin{tabular}{@{}*{4}{c@{\hspace{12pt}}}@{\hspace{-12pt}}}
    {\small DooD (QD)}
    & {\small DooD (OM)}
    & {\small AIR (OM)}
    & {\small GNS\cite{feinman2020learning} (OM)} \\
    \concropgt{char_con_gen/char_con_gen_dood-qd.pdf}
    & \concropgt{char_con_gen/char_con_gen_dood-om.pdf}
    & \concropgt{char_con_gen/char_con_gen_dair-om.pdf}
    & \concropgt{char_con_gen/char_con_gen_gns.pdf} \\
    \midrule
    \concropimg{char_con_gen/char_con_gen_dood-qd.pdf}
    & \concropimg{char_con_gen/char_con_gen_dood-om.pdf}
    & \concropimg{char_con_gen/char_con_gen_dair-om.pdf}
    & \concropimg{char_con_gen/char_con_gen_gns.pdf} \\
  \end{tabular}
  \caption{Given a target image of a handwritten Omniglot character, our model produces realistic new exemplars by inferring an explicit stroke-based representation.}
  \label{fig:char_con_gen}
\end{figure}

DooD generates realistic new exemplars of complex QuickDraw drawings and Omniglot characters (\cref{fig:char_con_gen}) thanks to the accurate inference and the ability to add noise to explicitly parametrised strokes.
While we can add an equivalent token model for AIR by (i) adding a Gaussian noise to the uninterpretable feature vector representing each part and (ii) applying a Gaussian affine transformation to the rendered image, the new exemplars are not as realistic both because of worse inference and the hard-to-control variations of the part vectors.
GNS generates realistic conditional samples, but notably still makes unnatural samples in multiple instances (e.g. in column 1, 2 the detachments of strokes)\footnote{Characters are different for GNS \cite{feinman2020learning} because those are the only publicly available samples for it.}, despite having a hierarchical model learned with multi-levels of supervision.

\begin{figure}[!t]
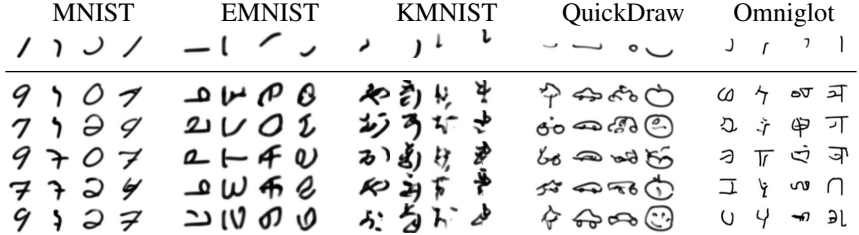

  \def\parcroptgt#1{\includegraphics[width=2.6ex, trim={0 {0.8\textheight} 0 0},clip]{#1}}
  \def\parcropimg#1{\includegraphics[width=2.6ex, trim={0 {0.3\textheight} 0 {0.1\textheight}},clip]{#1}}
  \centering
  \vspace*{\baselineskip}
  \begin{tabular}{@{}*{5}{*{4}{c@{\hspace{2pt}}}@{\hspace*{12pt}}}@{\hspace{-12pt}}}
    \multicolumn{4}{c}{MNIST}
    & \multicolumn{4}{c}{EMNIST}
    & \multicolumn{4}{c}{KMNIST}
    & \multicolumn{4}{c}{QuickDraw}
    & \multicolumn{4}{c}{Omniglot} \\
     \parcroptgt{partial_completion_seeded_ind3/MNIST-par_com-1}
    & \parcroptgt{partial_completion_seeded_ind3/MNIST-par_com-12}
    & \parcroptgt{partial_completion_seeded_ind3/MNIST-par_com-17}
    & \parcroptgt{partial_completion_seeded_ind3/MNIST-par_com-36}
    & \parcroptgt{partial_completion_seeded_ind3/EMNIST-par_com-0}
    & \parcroptgt{partial_completion_seeded_ind3/EMNIST-par_com-4}
    & \parcroptgt{partial_completion_seeded_ind3/EMNIST-par_com-18}
    & \parcroptgt{partial_completion_seeded_ind3/EMNIST-par_com-38}
    & \parcroptgt{partial_completion_seeded_ind3/KMNIST-par_com-1}
    & \parcroptgt{partial_completion_seeded_ind3/KMNIST-par_com-7}
    & \parcroptgt{partial_completion_seeded_ind3/KMNIST-par_com-10}
    & \parcroptgt{partial_completion_seeded_ind3/KMNIST-par_com-39}
    & \parcroptgt{partial_completion_seeded_ind3/Quickdraw-par_com-4}
    & \parcroptgt{partial_completion_seeded_ind3/Quickdraw-par_com-7}
    & \parcroptgt{partial_completion_seeded_ind3/Quickdraw-par_com-12}
    & \parcroptgt{partial_completion_seeded_ind3/Quickdraw-par_com-38}
    & \parcroptgt{partial_completion_seeded_ind3/Omniglot-par_com-2}
    & \parcroptgt{partial_completion_seeded_ind3/Omniglot-par_com-13}
    & \parcroptgt{partial_completion_seeded_ind3/Omniglot-par_com-24}
    & \parcroptgt{partial_completion_seeded_ind3/Omniglot-par_com-29} \\
    \midrule
     \parcropimg{partial_completion_seeded_ind3/MNIST-par_com-1}
    & \parcropimg{partial_completion_seeded_ind3/MNIST-par_com-12}
    & \parcropimg{partial_completion_seeded_ind3/MNIST-par_com-17}
    & \parcropimg{partial_completion_seeded_ind3/MNIST-par_com-36}
    & \parcropimg{partial_completion_seeded_ind3/EMNIST-par_com-0}
    & \parcropimg{partial_completion_seeded_ind3/EMNIST-par_com-4}
    & \parcropimg{partial_completion_seeded_ind3/EMNIST-par_com-18}
    & \parcropimg{partial_completion_seeded_ind3/EMNIST-par_com-38}
    & \parcropimg{partial_completion_seeded_ind3/KMNIST-par_com-1}
    & \parcropimg{partial_completion_seeded_ind3/KMNIST-par_com-7}
    & \parcropimg{partial_completion_seeded_ind3/KMNIST-par_com-10}
    & \parcropimg{partial_completion_seeded_ind3/KMNIST-par_com-39}
    & \parcropimg{partial_completion_seeded_ind3/Quickdraw-par_com-4}
    & \parcropimg{partial_completion_seeded_ind3/Quickdraw-par_com-7}
    & \parcropimg{partial_completion_seeded_ind3/Quickdraw-par_com-12}
    & \parcropimg{partial_completion_seeded_ind3/Quickdraw-par_com-38}
    & \parcropimg{partial_completion_seeded_ind3/Omniglot-par_com-2}
    & \parcropimg{partial_completion_seeded_ind3/Omniglot-par_com-13}
    & \parcropimg{partial_completion_seeded_ind3/Omniglot-par_com-24}
    & \parcropimg{partial_completion_seeded_ind3/Omniglot-par_com-29}
  \end{tabular}
  \caption{Given a partially drawn character, our model can generate a realistic distribution over its completions by sampling from the generative model conditioned on the image of the partial character.}
  \label{fig:partial_gen_new}
\end{figure}

\paragraph{Partial completion.}
As with inferring an entire figure in the previous case, we can interpret conditional generation in a slightly different way as well---where the condition is an \emph{initialization} of a number, character, or figure, and the model tries to extend/complete it as best it can (\cref{fig:partial_gen_new}).
To do this, we first employ the recognition model over the partial figure to compute the hidden states of the shared recurrent networks.
Next, starting with these computed states, we set the canvas-so-far~$x_{<t}$ to be the partial figure itself and then unroll the generative model from that point onwards.
As can be seen in the figure, DooD can generate a varied range of completions for each initial stroke, demonstrating its versatility and the utility of its learnt representations.

\paragraph{One-shot classification.}\label{para:one_shot_classification}

\begin{wraptable}[11]{r}{0.33\textwidth}
  \vspace*{-1.4\baselineskip}
  \caption{\small Accuracy in one-shot classification, without data augmentation (DA), extra stroke-data supervision (ES), or 2-way classification (2W).}
  \label{tab:one_shot_clf_new}
  \centering
  \scalebox{0.88}{%
  \begin{tabular}{@{}l@{\hspace{1ex}}*{4}{c@{\hspace{1ex}}}@{\hspace{-1ex}}}
    \toprule
    Model
    & DA & ES & 2W & Accuracy\\
    \midrule
    AIR
    & \xmark & \xmark & \xmark & 14.5\% \\
    DooD
    & \xmark & \xmark & \xmark & 73.5\% \\
    \cmidrule(lr){1-5}
    VHE\cite{hewitt2018variational}
    & \cmark & \cmark & \xmark & 81.3\% \\
    GNS\cite{feinman2020learning}
    & \xmark & \cmark & \cmark & 94.3\% \\
    BPL\cite{lake2015human}
    & \xmark & \cmark & \cmark & 96.7\% \\
    \bottomrule
  \end{tabular}}
\end{wraptable}
Finally, we can apply the type-token hierarchical generative model used for generating new exemplars to perform within-alphabet, 20-way one-shot classification.
The key quantity needed for performing this task is the posterior predictive score of the query image $x^{(T)}$ given a set of support images $\{x^{(c)}\}_c$, $\hat{c} = \arg\max_c p(x^{(T)} | x^{(c)})$, which requires marginalizing over the token variables corresponding to $x^{(T)}$ and $x^{(c)}$, and the type variable of $x^{(c)}$.
Following \citep{feinman2020learning, lake2015human}, we approximate this score by sampling from the recognition model given $x^{(c)}$, and perform gradient-based optimization to marginalize out the token variable of $x^{(T)}$ (details in \cref{app:one_shot_classification_details}).
We find that DooD outperforms the neural baseline (AIR), while attaining a competitive accuracy in comparison to other baselines (\cref{tab:one_shot_clf_new}) without requiring additional forms of support such as data augmentation, supervision for strokes, or more complex ways of computing the accuracy such as two-way scoring (predicting $\hat{c}=\arg\max_c \frac{p(x^{(c)} | x^{(T)})}{p(x^{(c)})} p(x^{(T)} | x^{(c)})$).

\section{Related Work}

Our work takes inspiration from \citet{lake2015human}'s symbolic generative modelling approach which hypothesises that the human ability to generalise comes from our causal and compositional understanding that characters are generated by composing substrokes into strokes, strokes into characters and rendering characters to images.
As a result, \citet{lake2015human} demonstrate human-like generalization on a wide range of tasks.
However, it is trained using stroke sequences, and inference is performed using expensive Markov chain Monte Carlo sampling.

We combine features of neuro-symbolic generative models and deep generative models to be able to generalise well across tasks while using amortised inference and being unsupervised.
From neuro-symbolic models, we share key features of \citet{feinman2020learning}'s model like (i) using the canvas-so-far in the generative model and adopt a similar feature in the recognition model like \citet{ellis2019write}, (ii) parametrizing parts as splines and using a differentiable spline renderer, (iii) extending the model to have a type-token hierarchy for generating new exemplars and performing one-shot classification.
Our model can be seen as an extension of \citep{feinman2020learning} that learns directly from images and uses a recognition model for amortised inference.
Like \citet{hewitt2020learning}, we learn how to infer a stroke sequence directly from images using a differentiable renderer but infer strokes directly instead of learning a stroke bank and use a more flexible parametrization of strokes based on a differentiable spline renderer, leading to a more accurate model.

Similar to deep generative modelling approaches like \citep{gregor2015draw,eslami2016attend,rezende2016one}, we use attention to focus on parts of the canvas we want to generate to or recognise from which allows neural networks to learn simpler and hence more generalizable mappings.
To be able to train our model from unsupervised images, we adopt the NVIL control variate \citep{mnih2014neural} used by \citet{eslami2016attend} to be able to train a model with a discrete stop-drawing latent variable.
This family of models, along with deep meta-learning approaches \citep{vinyals2016matching,finn2017model,snell2017prototypical}, is easier to learn due to the lack of symbolic variables and results in a fast amortised recognition model.
However, the lack of strong inductive biases leads to poor and unreliable generalization \citep{lake2019omniglot}.
We also share idea with other works combining deep learning and explicit stroke modelling \citep{ganin2018synthesizing,mellor2019unsupervised,ha2017neural,aksan2020cose}, but we focus on learning a principled generative model which allows tackling tasks like one-shot classification and generating new exemplars, in addition to conditional and unconditional sampling.

\section{Conclusion}

We demonstrated that DooD generalises across datasets and across tasks thanks to an explicit symbolic parametrization of strokes and guided execution.
This allows us to train on one dataset such as MNIST and generalise to a more complex, out-of-distribution dataset such as Omniglot.
Given a compositional representation and an associated learned sequential prior, DooD can be applied to additional tasks in the Omniglot challenge like generating new exemplars and one-shot classification by extending it to have a type-token hierarchy.
Our model produces realistic new exemplars without blur and artefacts unlike deep generative models.

More broadly, DooD is an example of a system that successfully combines symbolic generative models to achieve generalization and deep learning models to handle raw perceptual data and perform fast amortised inference while being learned from unsupervised data.
We believe these principles can be useful for building fast, reliable and robust learning systems going beyond stroke-based data.

\bibliographystyle{abbrvnat}
\bibliography{main}

\newpage

\appendix

\begin{center}
\textbf{\large Supplement to: Drawing out of Distribution with\\ Neuro-Symbolic Generative Models}
\end{center}

\section{Dataset Details}
\label{app:dataset}
All dataset images are scaled to 50x50 in grayscale, with dataset-specific configuration list below.
\begin{description}
\item[MNIST, KMNIST:] we use the original split of with 60k images for training, 10k for tests. Each image belongs to 1 of the 10 classes.
\item[EMNIST:] we use the ``balanced'' split of the dataset with 112,800 training images and 18,800 testing images, separated into 47 classes.
\item[QuickDraw:] we use the 10-category version of dataset, where each has 4k/1k training/testing samples, as found on \url{https://github.com/XJay18/QuickDraw-pytorch}.
\item[Omniglot:] we use the original split~\cite{lake2015human}, with inverted black and white pixels. For one-shot classification (\cref{para:one_shot_classification}), we use the original task-split, as found on \url{https://github.com/brendenlake/omniglot}. It has 20 episodes, each a 20-way, 1-shot, within-alphabet classification task.
\end{description}

\section{Model Details}
\label{app:model}
\subsection{Differentiable Renderer}
\label{app:renderer}
The renderer outputs grayscale images when inputs $\tilde{\zwhat}_t$ that are interpreted as control-point coordinates. It has three parameters $(\sigma^t, \sslope^t, \gslope)$, where the first two are per-stroke and $\gslope$ is per-character. It renders a stroke through two steps: 1) compute a sample curve based on the control points; 2) rasterize the output image from the sample curve.

With $J$ control points, $\tilde{\zwhat}_t = [p^t_j]_{j=0}^{J-1}$, each with their $x, y$ coordinates, a sample curve with $S$ sample points can be computed using the explicit definition of Bézier curves, where $n$ is one of the $S$ numbers ranging $[0, 1]$:

\begin{align}
  \mathbf{b}^t_n
  = \sum_{j=0}^{J-1} {(J-1)\choose j} (1-n)^{J-1-j} n^j p_j
\end{align}

In our case, we use $S=100$ samples spaced evenly between $[0, 1]$.

With $S$ points $[\mathbf{b}^t_n]$, an image $\vardbtilde{x}^t$ can be rasterized, as indexed by $h \in [0, H), w\in [0, W)$, where $H, W$ are image dimensions, and with the pixel intensity $\vardbtilde{x}^t_{hw}$ given by:

\begin{align}
  \vardbtilde{x}^t_{hw}
  = \sum_n \frac{(h - \mathbf{b}^t_{n,x})^2 (w - \mathbf{b}^t_{n,y})^2} {(\sigma^t)^2}
\end{align}

where $\mathbf{b}^t_{n, x}, \mathbf{b}^t_{n,y}$ stands for the $x, y$ coordinates of the sample point $\mathbf{b}^t_n$, and $\sigma^t$ is the renderer parameter roughly in control of the blur of the rendering output.

As an effect of this rasterizing procedure, the pixel intensity can be arbitrarily large. To normalize it to be always inside $[0, 1]$, we apply a max-normalization followed by a parametrized tanh function:

\begin{align}
  \vardbtilde{x}^{t'}
  &= \mathrm{maxnorm} (\vardbtilde{x})
    = \frac{\vardbtilde{x}}{\mathrm{max}(\vardbtilde{x})} \\
  x^t_{hw}
  &= \mathrm{normalize\_stroke} (\vardbtilde{x}^{t'}_{hw};\sslope^t)
    = \tanh(\frac{\vardbtilde{x}^{t'}_{hw}}{\sslope^t})
\end{align}
The max-normalization divides each image's pixel values by the highest pixel value of that image, normalizing all pixels to the range $[0, 1]$.
Both steps here are important because with just the max-norm, the maximum pixel value of each image is always 1, which is usually not preferred. Conversely, with just the tanh-normalization, $\sslope^t$ would be required to vary in a much greater range for the output image to look as intended, as $\mathrm{max}(\vardbtilde{x})$ can range from tens to thousands.

With the image pixel value of an individual stroke being in $[0,1]$, an element-wise sum of all strokes' rendering could still be larger than 1. Hence another parametrized tanh function is used to get the canvas-so-far at time $T'$:

\begin{align}
  x^{\leq T'}_{hw}
  = \mathrm{normalize\_canvas} ([x^t_{hw}]_{t=1}^{T'};\gslope)
  = \tanh(\frac{\sum_{t=1}^{T'} x^t_{hw}}{\gslope})
\end{align}

\subsection{Neural Network Configurations}
\label{app:nn_config}
DooD and AIR in our experiments share the overall neural components.

Convolutional neural nets (CNN) are used as feature extractors for images (one for canvas-so-far, attention window, target, another for residual and its counterparts). Each CNN is equipped 2 Conv2d layers with 3x3 kernel and stride 1 followed by a 1-layer MLP that outputs 256-dim features. The Conv2d layers goes from 1 to 8, then to 16 channels. Notably, we don't use any Max Pooling layer to avoid the spatial-invariant property.

All the prior, posterior distribution parameters are output by their respective MLP (results in 6 separate MLPs). Despite varying input, output dimensions, they share the main architecture: 2 256-dim  hidden-layers with tanh non-linearity. The renderer parameters as detailed in \cref{app:renderer} are predicted by another MLP of the same form, but not modelled as latent variables in our implementation.
As a result, DooD employs 7 MLPs.
For AIR, An MLP is used as the decoder (i.e., renderer) for AIR, with the same configurations as above.

On top of these, GRUs\citep{chung2015recurrent} with 256-dim hidden states are employed for the layout and stroke RNNs.

\subsection{Token Model}
\label{app:token_model}
To fit our model naturally into the hierarchical Bayesian formulation of the character-conditioned generation and the one-shot classification task, we inserted a plug-and-play \textbf{token model} to our generative model.
With the learned generative and recognition model, we can regard the learned \textbf{prior} $p(\zpres,\zwhere,\zwhat)$ as a high level \textbf{type model} $p(\psi)$ and incorporate a token model $p(z|\psi)$, where $\psi, z$ denote $(\zpres,\zwhere,\zwhat)$, $(\zpres,\zwhere,\zwhat')$, respectively (only potentially different in $\zwhat$ vs. $\zwhat'$). We can then consider the learned variational posterior $q(\zpres,\zwhere,\zwhat|x)$ to be directly on the type variable $\psi$, i.e. $q(\psi|x)$.
The token model $p(z|\psi)$ captures the plausible structural variability of various instances of a character (including affine transformations, motor noise; all embodied in $\zwhat'$ given $\zwhat$).
This can either be learned or set by heuristics.\\
In our experiment, we simply leverage a uniform distribution over a range of affine transformations and a spherical normal distribution for motor-noise.
Note that the flexibility of doing this is thanks to the symbolic latent representation that DooD has, while models with distributed latent representations lack.
In detail, the motor noise model has a standard Gaussian distribution with mean centered on the control points and scale $1e-3$. The affine model uses uniform distribution and has $x, y$ shift value ranging $[-.2,.2]$, $x, y$ scale $[.8, 1.2]$, rotation $[-.25\pi, .25\pi]$, $x, y$ shear $[-.25\pi, .25\pi]$

\newpage
\section{Training Details}
\label{app:training}
The model is trained with the Adam \cite{kingma2014adam} optimizer with a learning rate of 1e-3 for the parameters whose gradients are estimated with NVIL \cite{mnih2014neural} and 1e-4 for the rest, neither with weight decay. The intermediate canvas-so-far $x_{\leq t}$ for $t\neq T$ and residual $\Delta x_t$ produced at each step are detached for both DooD and AIR from the gradient graph for training stability, effectively making them not act as a medium for backpropagation-through-time.
For DooD and it's ablations, $\beta=4$ is used in the loss term, whereas $\beta=5$ is used on AIR, both tuned on the MNIST-trained across-dataset generalization task.

The initial parameters of the last layer of the $\zwhere$ MLPs are set to predict identity transformations as per \cite{jaderberg2015spatial}. For the Omniglot dataset only, the initial weights for $\zpres$ MLP's last layer is zeroed and the initial bias is set to a high value (e.g. 8) before passing through a sigmoid function for normalization, because otherwise the model would quickly go to using no steps due to the greater difficulty in joint learning and inference on Omniglot.

\subsection{Stochastic Gradient Estimators}
\label{app:grad_est}
One way of learning the parameters of the generative model $\theta$ and the inference network $\phi$ is by jointly maximizing the lower bound of the marginal likelihood of an image $x$, denoting the joint latent variables by $z$:

\begin{align}
    \log \ptheta(x)
    &= \log\int dz \ptheta(x, z)
    = \log\int dz \qphi(z|x) \frac{\ptheta(x, z)}{\qphi(z|x)} \nonumber \\
    &= \log\E_{\qphi}\left[ \frac{\ptheta(x, z)}{\qphi(z|x)} \right]
    \geq \E_{\qphi}\left[\log\frac{ \ptheta(x, z)} {\qphi(z | x)}\right] \nonumber \\
    &= \E_{\qphi}[\log\ptheta(z)]
      + \E_{\qphi}[\log\ptheta(x|z)]
      - \E_{\qphi}[\log\qphi(z|x)]
      =: \L(\theta, \phi)
\end{align}

A Monte Carlo gradient estimator for $\pdv{}{\theta}\L$ is relatively easy to compute by drawing $z \sim \qphi(\cdot|x)$ and computing $ \pdv{}{\theta}  \log p_\theta(x, z)$ as the model is differentiable w.r.t. its parameters.

Estimating the gradient for $\pdv{}{\phi}\L$ is more involved as the parameters $\phi$ are also used when drawing samples from  $\qphi$. To address this, for each step $t$, denote $\omega^t$ all the parameters of the distribution on variables at $t$, $z^t$. 
The full gradient can therefore be obtained via chain rule: $\pdv{\L}{\phi}=\sum_t\pdv{\L}{\omega^t}\pdv{\omega^t}{\phi}$.\\
Define $\ll := \log \frac{\ptheta(x, z)}{\qphi(z|x)}$, we can write the loss as $\L(\theta,\phi) = \E_{\qphi}[\ll]$, and let $z^t$ be either the continuous or the discrete subset of latent variables in $(\zwhere^t, \zwhat^t, \zpres^t)$. How to proceed with computing $\pdv{\L}{\omega^t}$ depends on whether $z^t$ is discrete or continuous.

\textbf{Continuous.} For continuous random variable $z^t$, we can use the reparametrization trick to back-propagate through $z^t$ \cite{kingma2013auto, schulman2015gradient}. The trick suggest that for many continuous random variables, drawing a sample $z^t$ from the distribution parametrized by $\omega^t$ yields an equivalent result as taking the output of a deterministic function inputting some random noise variable $\xi$ and parameter $\omega^t$, $z^t = f(\xi, \omega^t)$ where $\xi$ is sampled from some fixed noise distribution $p(\xi)$. This results in the estimator: $\pdv{\L}{\omega^t} \approx \pdv{\ll}{z^t} \pdv{f}{\omega^t}$.

\textbf{Discrete.} For discrete variables such as $\zpres^t$, the reparametrization trick can't be applied. Instead, we resort to the REINFORCE estimator \cite{mnih2014neural, schulman2015gradient}, with a Monte Carlo estimate of the gradient: $\pdv{\L}{\omega^t} \approx \pdv{\log\qphi(z|x)}{\omega^t} \ll$.

This can be derived as follows (denote $\ll$ by $\ell(z)$ and $\qphi(z|x)$ by $\qphi(z)$ to simplify notation):

\begin{align}
    \pdv{\L}{\omega^t}
    &= \pdv{}{\omega^t} \int  \qphi(z) \ell(z) dz \nonumber\\
    &= \int\left(\pdv{}{\omega^t}\log \qphi(z)\right)  \qphi(z) \ell(z) dz \nonumber\\
    &= \E_{\qphi(z)} \left[\pdv{}{\omega^t} \log \qphi(z) \ell(z)\right] \nonumber\\
    &\approx \pdv{\log\qphi(z)}{\omega^t} \ell(z)
\end{align}

This basic form usually results in a high variance, and we can significantly reduce it by using only local learning signals and a structured neural baseline\cite{mnih2014neural}.  The former suggests that we can remove the terms in $\ell(z)$ that don't depend on $\omega^t$ without affecting the result, this allows us to substitute $\ell(z)$ with $\ell^t(z) := \log\ptheta(x|z) \ptheta(z^{t:T}) / \qphi({z^{t:T}})$ such that it only uses learning signals dependent on $\omega^t$.  The latter suggests subtracting a control variate $b(z^{<t},x)$, which takes in $x$ and the previous variables $z^{<t}$ detached from the gradient graph, from $\ell^t(\cdot)$. It is trained by minimizing the mean squared error between $\ell^t(\cdot)$ and $b(z^{<t},x)$, i.e., $\L_b := \E_{\qphi}[(\ell^t(z)-b(z^{<t},x))^2]$. This yields an lower-variance estimator used in learning $\pdv{\L}{\omega^t}\approx\pdv{\qphi(z^t)}{\omega^t} \left( \ell^t(z) - b(z^{<t}, x) \right)$.  Finally, the learning signal, $\left( \ell^t(z) - b(z^{<t}, x) \right)$, is centered and smoothed as in \citep{mnih2014neural}. And the final loss function can be written as $\hat\L = \L + \L_b$.

\section{Evaluation Details}
\label{app:evaluation}
\subsection{More across-dataset generalization results}
\cref{fig:mnist_generalization} demonstrates each model's performance when trained on MNIST. Here we show an instance of \cref{fig:mnist_generalization} with additional results for models trained on each of the 5 datasets (except for the baseline that didn't work on that particular dataset). 
We further present a \cref{fig:cross_ds_mll_eval}-style confusion matrix for DooD-EG.

\newpage
\subsubsection{EMNIST-trained models}
\begin{center}
  \begin{sideways}
    \begin{minipage}[c]{0.96\textheight}
    \scalebox{0.7}{\def\targetimg#1{\includegraphics[width=0.03\textwidth,trim={0 {0.6\textheight} 0 0},clip]{#1}}
\def\reconimg#1{\includegraphics[width=0.03\textwidth,trim={0 0 0 {0.097\textheight}},clip]{#1}}
\setlength\tabcolsep{1pt}
\begin{tabular}{@{}c@{\quad}*{8}{c}@{\hspace{20pt}}*{4}{*{8}{c}@{\hspace{10pt}}}@{\hspace{-10pt}}}
  (a)
  & \multicolumn{8}{c}{\quad EMNIST \quad\(\bm{\to}\)\,}
  & \multicolumn{8}{c}{MNIST}
  & \multicolumn{8}{c}{KMNIST}
  & \multicolumn{8}{c}{QuickDraw}
  & \multicolumn{8}{c}{Omniglot} \\
  \midrule
  $x$
  & \targetimg{supp_cross_recon/target_EMNIST_5.pdf}
  & \targetimg{supp_cross_recon/target_EMNIST_6.pdf}
  & \targetimg{supp_cross_recon/target_EMNIST_7.pdf}
  & \targetimg{supp_cross_recon/target_EMNIST_8.pdf}
  & \targetimg{supp_cross_recon/target_EMNIST_9.pdf}
  & \targetimg{supp_cross_recon/target_EMNIST_10.pdf}
  & \targetimg{supp_cross_recon/target_EMNIST_11.pdf}
  & \targetimg{supp_cross_recon/target_EMNIST_12.pdf}
  & \targetimg{supp_cross_recon/target_MNIST_0.pdf}
  & \targetimg{supp_cross_recon/target_MNIST_1.pdf}
  & \targetimg{supp_cross_recon/target_MNIST_2.pdf}
  & \targetimg{supp_cross_recon/target_MNIST_3.pdf}
  & \targetimg{supp_cross_recon/target_MNIST_4.pdf}
  & \targetimg{supp_cross_recon/target_MNIST_5.pdf}
  & \targetimg{supp_cross_recon/target_MNIST_6.pdf}
  & \targetimg{supp_cross_recon/target_MNIST_7.pdf}
  & \targetimg{supp_cross_recon/target_KMNIST_5.pdf}
  & \targetimg{supp_cross_recon/target_KMNIST_6.pdf}
  & \targetimg{supp_cross_recon/target_KMNIST_7.pdf}
  & \targetimg{supp_cross_recon/target_KMNIST_8.pdf}
  & \targetimg{supp_cross_recon/target_KMNIST_9.pdf}
  & \targetimg{supp_cross_recon/target_KMNIST_13.pdf}
  & \targetimg{supp_cross_recon/target_KMNIST_14.pdf}
  & \targetimg{supp_cross_recon/target_KMNIST_15.pdf}
  & \targetimg{supp_cross_recon/target_Quickdraw_5.pdf}
  & \targetimg{supp_cross_recon/target_Quickdraw_6.pdf}
  & \targetimg{supp_cross_recon/target_Quickdraw_7.pdf}
  & \targetimg{supp_cross_recon/target_Quickdraw_8.pdf}
  & \targetimg{supp_cross_recon/target_Quickdraw_9.pdf}
  & \targetimg{supp_cross_recon/target_Quickdraw_10.pdf}
  & \targetimg{supp_cross_recon/target_Quickdraw_11.pdf}
  & \targetimg{supp_cross_recon/target_Quickdraw_12.pdf}
  & \targetimg{supp_cross_recon/target_Omniglot_5.pdf}
  & \targetimg{supp_cross_recon/target_Omniglot_6.pdf}
  & \targetimg{supp_cross_recon/target_Omniglot_7.pdf}
  & \targetimg{supp_cross_recon/target_Omniglot_8.pdf}
  & \targetimg{supp_cross_recon/target_Omniglot_9.pdf}
  & \targetimg{supp_cross_recon/target_Omniglot_0.pdf}
  & \targetimg{supp_cross_recon/target_Omniglot_1.pdf}
  & \targetimg{supp_cross_recon/target_Omniglot_2.pdf}
  \\
  \midrule
  \raisebox{8ex}{\rotatebox[origin=c]{90}{DooD}}
  & \reconimg{supp_cross_recon/dood-em_EMNIST_5.pdf}
  & \reconimg{supp_cross_recon/dood-em_EMNIST_6.pdf}
  & \reconimg{supp_cross_recon/dood-em_EMNIST_7.pdf}
  & \reconimg{supp_cross_recon/dood-em_EMNIST_8.pdf}
  & \reconimg{supp_cross_recon/dood-em_EMNIST_9.pdf}
  & \reconimg{supp_cross_recon/dood-em_EMNIST_10.pdf}
  & \reconimg{supp_cross_recon/dood-em_EMNIST_11.pdf}
  & \reconimg{supp_cross_recon/dood-em_EMNIST_12.pdf}
  & \reconimg{supp_cross_recon/dood-em_MNIST_0.pdf}
  & \reconimg{supp_cross_recon/dood-em_MNIST_1.pdf}
  & \reconimg{supp_cross_recon/dood-em_MNIST_2.pdf}
  & \reconimg{supp_cross_recon/dood-em_MNIST_3.pdf}
  & \reconimg{supp_cross_recon/dood-em_MNIST_4.pdf}
  & \reconimg{supp_cross_recon/dood-em_MNIST_5.pdf}
  & \reconimg{supp_cross_recon/dood-em_MNIST_6.pdf}
  & \reconimg{supp_cross_recon/dood-em_MNIST_7.pdf}
  & \reconimg{supp_cross_recon/dood-em_KMNIST_5.pdf}
  & \reconimg{supp_cross_recon/dood-em_KMNIST_6.pdf}
  & \reconimg{supp_cross_recon/dood-em_KMNIST_7.pdf}
  & \reconimg{supp_cross_recon/dood-em_KMNIST_8.pdf}
  & \reconimg{supp_cross_recon/dood-em_KMNIST_9.pdf}
  & \reconimg{supp_cross_recon/dood-em_KMNIST_13.pdf}
  & \reconimg{supp_cross_recon/dood-em_KMNIST_14.pdf}
  & \reconimg{supp_cross_recon/dood-em_KMNIST_15.pdf}
  & \reconimg{supp_cross_recon/dood-em_Quickdraw_5.pdf}
  & \reconimg{supp_cross_recon/dood-em_Quickdraw_6.pdf}
  & \reconimg{supp_cross_recon/dood-em_Quickdraw_7.pdf}
  & \reconimg{supp_cross_recon/dood-em_Quickdraw_8.pdf}
  & \reconimg{supp_cross_recon/dood-em_Quickdraw_9.pdf}
  & \reconimg{supp_cross_recon/dood-em_Quickdraw_10.pdf}
  & \reconimg{supp_cross_recon/dood-em_Quickdraw_11.pdf}
  & \reconimg{supp_cross_recon/dood-em_Quickdraw_12.pdf}
  & \reconimg{supp_cross_recon/dood-em_Omniglot_5.pdf}
  & \reconimg{supp_cross_recon/dood-em_Omniglot_6.pdf}
  & \reconimg{supp_cross_recon/dood-em_Omniglot_7.pdf}
  & \reconimg{supp_cross_recon/dood-em_Omniglot_8.pdf}
  & \reconimg{supp_cross_recon/dood-em_Omniglot_9.pdf}
  & \reconimg{supp_cross_recon/dood-em_Omniglot_0.pdf}
  & \reconimg{supp_cross_recon/dood-em_Omniglot_1.pdf}
  & \reconimg{supp_cross_recon/dood-em_Omniglot_2.pdf}
  \\
  \midrule
  \raisebox{8ex}{\rotatebox[origin=c]{90}{AIR}}
  & \reconimg{supp_cross_recon/dair-em_EMNIST_5.pdf}
  & \reconimg{supp_cross_recon/dair-em_EMNIST_6.pdf}
  & \reconimg{supp_cross_recon/dair-em_EMNIST_7.pdf}
  & \reconimg{supp_cross_recon/dair-em_EMNIST_8.pdf}
  & \reconimg{supp_cross_recon/dair-em_EMNIST_9.pdf}
  & \reconimg{supp_cross_recon/dair-em_EMNIST_10.pdf}
  & \reconimg{supp_cross_recon/dair-em_EMNIST_11.pdf}
  & \reconimg{supp_cross_recon/dair-em_EMNIST_12.pdf}
  & \reconimg{supp_cross_recon/dair-em_MNIST_0.pdf}
  & \reconimg{supp_cross_recon/dair-em_MNIST_1.pdf}
  & \reconimg{supp_cross_recon/dair-em_MNIST_2.pdf}
  & \reconimg{supp_cross_recon/dair-em_MNIST_3.pdf}
  & \reconimg{supp_cross_recon/dair-em_MNIST_4.pdf}
  & \reconimg{supp_cross_recon/dair-em_MNIST_5.pdf}
  & \reconimg{supp_cross_recon/dair-em_MNIST_6.pdf}
  & \reconimg{supp_cross_recon/dair-em_MNIST_7.pdf}
  & \reconimg{supp_cross_recon/dair-em_KMNIST_5.pdf}
  & \reconimg{supp_cross_recon/dair-em_KMNIST_6.pdf}
  & \reconimg{supp_cross_recon/dair-em_KMNIST_7.pdf}
  & \reconimg{supp_cross_recon/dair-em_KMNIST_8.pdf}
  & \reconimg{supp_cross_recon/dair-em_KMNIST_9.pdf}
  & \reconimg{supp_cross_recon/dair-em_KMNIST_13.pdf}
  & \reconimg{supp_cross_recon/dair-em_KMNIST_14.pdf}
  & \reconimg{supp_cross_recon/dair-em_KMNIST_15.pdf}
  & \reconimg{supp_cross_recon/dair-em_Quickdraw_5.pdf}
  & \reconimg{supp_cross_recon/dair-em_Quickdraw_6.pdf}
  & \reconimg{supp_cross_recon/dair-em_Quickdraw_7.pdf}
  & \reconimg{supp_cross_recon/dair-em_Quickdraw_8.pdf}
  & \reconimg{supp_cross_recon/dair-em_Quickdraw_9.pdf}
  & \reconimg{supp_cross_recon/dair-em_Quickdraw_10.pdf}
  & \reconimg{supp_cross_recon/dair-em_Quickdraw_11.pdf}
  & \reconimg{supp_cross_recon/dair-em_Quickdraw_12.pdf}
  & \reconimg{supp_cross_recon/dair-em_Omniglot_5.pdf}
  & \reconimg{supp_cross_recon/dair-em_Omniglot_6.pdf}
  & \reconimg{supp_cross_recon/dair-em_Omniglot_7.pdf}
  & \reconimg{supp_cross_recon/dair-em_Omniglot_8.pdf}
  & \reconimg{supp_cross_recon/dair-em_Omniglot_9.pdf}
  & \reconimg{supp_cross_recon/dair-em_Omniglot_0.pdf}
  & \reconimg{supp_cross_recon/dair-em_Omniglot_1.pdf}
  & \reconimg{supp_cross_recon/dair-em_Omniglot_2.pdf}
  \\
  \midrule
  \raisebox{8ex}{\rotatebox[origin=c]{90}{DooD-EG}}
  & \reconimg{supp_cross_recon/dood_eg-em_EMNIST_5.pdf}
  & \reconimg{supp_cross_recon/dood_eg-em_EMNIST_6.pdf}
  & \reconimg{supp_cross_recon/dood_eg-em_EMNIST_7.pdf}
  & \reconimg{supp_cross_recon/dood_eg-em_EMNIST_8.pdf}
  & \reconimg{supp_cross_recon/dood_eg-em_EMNIST_9.pdf}
  & \reconimg{supp_cross_recon/dood_eg-em_EMNIST_10.pdf}
  & \reconimg{supp_cross_recon/dood_eg-em_EMNIST_11.pdf}
  & \reconimg{supp_cross_recon/dood_eg-em_EMNIST_12.pdf}
  & \reconimg{supp_cross_recon/dood_eg-em_MNIST_0.pdf}
  & \reconimg{supp_cross_recon/dood_eg-em_MNIST_1.pdf}
  & \reconimg{supp_cross_recon/dood_eg-em_MNIST_2.pdf}
  & \reconimg{supp_cross_recon/dood_eg-em_MNIST_3.pdf}
  & \reconimg{supp_cross_recon/dood_eg-em_MNIST_4.pdf}
  & \reconimg{supp_cross_recon/dood_eg-em_MNIST_5.pdf}
  & \reconimg{supp_cross_recon/dood_eg-em_MNIST_6.pdf}
  & \reconimg{supp_cross_recon/dood_eg-em_MNIST_7.pdf}
  & \reconimg{supp_cross_recon/dood_eg-em_KMNIST_5.pdf}
  & \reconimg{supp_cross_recon/dood_eg-em_KMNIST_6.pdf}
  & \reconimg{supp_cross_recon/dood_eg-em_KMNIST_7.pdf}
  & \reconimg{supp_cross_recon/dood_eg-em_KMNIST_8.pdf}
  & \reconimg{supp_cross_recon/dood_eg-em_KMNIST_9.pdf}
  & \reconimg{supp_cross_recon/dood_eg-em_KMNIST_13.pdf}
  & \reconimg{supp_cross_recon/dood_eg-em_KMNIST_14.pdf}
  & \reconimg{supp_cross_recon/dood_eg-em_KMNIST_15.pdf}
  & \reconimg{supp_cross_recon/dood_eg-em_Quickdraw_5.pdf}
  & \reconimg{supp_cross_recon/dood_eg-em_Quickdraw_6.pdf}
  & \reconimg{supp_cross_recon/dood_eg-em_Quickdraw_7.pdf}
  & \reconimg{supp_cross_recon/dood_eg-em_Quickdraw_8.pdf}
  & \reconimg{supp_cross_recon/dood_eg-em_Quickdraw_9.pdf}
  & \reconimg{supp_cross_recon/dood_eg-em_Quickdraw_10.pdf}
  & \reconimg{supp_cross_recon/dood_eg-em_Quickdraw_11.pdf}
  & \reconimg{supp_cross_recon/dood_eg-em_Quickdraw_12.pdf}
  & \reconimg{supp_cross_recon/dood_eg-em_Omniglot_5.pdf}
  & \reconimg{supp_cross_recon/dood_eg-em_Omniglot_6.pdf}
  & \reconimg{supp_cross_recon/dood_eg-em_Omniglot_7.pdf}
  & \reconimg{supp_cross_recon/dood_eg-em_Omniglot_8.pdf}
  & \reconimg{supp_cross_recon/dood_eg-em_Omniglot_9.pdf}
  & \reconimg{supp_cross_recon/dood_eg-em_Omniglot_0.pdf}
  & \reconimg{supp_cross_recon/dood_eg-em_Omniglot_1.pdf}
  & \reconimg{supp_cross_recon/dood_eg-em_Omniglot_2.pdf}
  \\
\end{tabular}

}
    \captionof{figure}{EMNIST-trained model generalize to other datasets.}
    \label{fig:emnist_generalization}
    \end{minipage}
  \end{sideways}
\end{center}

\newpage
\subsubsection{KMNIST-trained models}
\begin{center}
  \begin{sideways}
    \begin{minipage}[c]{0.96\textheight}
    \scalebox{0.7}{\def\targetimg#1{\includegraphics[width=0.03\textwidth,trim={0 {0.6\textheight} 0 0},clip]{#1}}
\def\reconimg#1{\includegraphics[width=0.03\textwidth,trim={0 0 0 {0.097\textheight}},clip]{#1}}
\setlength\tabcolsep{1pt}
\begin{tabular}{@{}c@{\quad}*{8}{c}@{\hspace{20pt}}*{4}{*{8}{c}@{\hspace{10pt}}}@{\hspace{-10pt}}}
  (a)
  & \multicolumn{8}{c}{\quad KMNIST \quad\(\bm{\to}\)\,}
  & \multicolumn{8}{c}{MNIST}
  & \multicolumn{8}{c}{EMNIST}
  & \multicolumn{8}{c}{QuickDraw}
  & \multicolumn{8}{c}{Omniglot} \\
  \midrule
  $x$
  & \targetimg{supp_cross_recon/target_KMNIST_5.pdf}
  & \targetimg{supp_cross_recon/target_KMNIST_6.pdf}
  & \targetimg{supp_cross_recon/target_KMNIST_7.pdf}
  & \targetimg{supp_cross_recon/target_KMNIST_8.pdf}
  & \targetimg{supp_cross_recon/target_KMNIST_9.pdf}
  & \targetimg{supp_cross_recon/target_KMNIST_13.pdf}
  & \targetimg{supp_cross_recon/target_KMNIST_14.pdf}
  & \targetimg{supp_cross_recon/target_KMNIST_15.pdf}
  & \targetimg{supp_cross_recon/target_MNIST_0.pdf}
  & \targetimg{supp_cross_recon/target_MNIST_1.pdf}
  & \targetimg{supp_cross_recon/target_MNIST_2.pdf}
  & \targetimg{supp_cross_recon/target_MNIST_3.pdf}
  & \targetimg{supp_cross_recon/target_MNIST_4.pdf}
  & \targetimg{supp_cross_recon/target_MNIST_5.pdf}
  & \targetimg{supp_cross_recon/target_MNIST_6.pdf}
  & \targetimg{supp_cross_recon/target_MNIST_7.pdf}
  & \targetimg{supp_cross_recon/target_EMNIST_5.pdf}
  & \targetimg{supp_cross_recon/target_EMNIST_6.pdf}
  & \targetimg{supp_cross_recon/target_EMNIST_7.pdf}
  & \targetimg{supp_cross_recon/target_EMNIST_8.pdf}
  & \targetimg{supp_cross_recon/target_EMNIST_9.pdf}
  & \targetimg{supp_cross_recon/target_EMNIST_10.pdf}
  & \targetimg{supp_cross_recon/target_EMNIST_11.pdf}
  & \targetimg{supp_cross_recon/target_EMNIST_12.pdf}
  & \targetimg{supp_cross_recon/target_Quickdraw_5.pdf}
  & \targetimg{supp_cross_recon/target_Quickdraw_6.pdf}
  & \targetimg{supp_cross_recon/target_Quickdraw_7.pdf}
  & \targetimg{supp_cross_recon/target_Quickdraw_8.pdf}
  & \targetimg{supp_cross_recon/target_Quickdraw_9.pdf}
  & \targetimg{supp_cross_recon/target_Quickdraw_10.pdf}
  & \targetimg{supp_cross_recon/target_Quickdraw_11.pdf}
  & \targetimg{supp_cross_recon/target_Quickdraw_12.pdf}
  & \targetimg{supp_cross_recon/target_Omniglot_5.pdf}
  & \targetimg{supp_cross_recon/target_Omniglot_6.pdf}
  & \targetimg{supp_cross_recon/target_Omniglot_7.pdf}
  & \targetimg{supp_cross_recon/target_Omniglot_8.pdf}
  & \targetimg{supp_cross_recon/target_Omniglot_9.pdf}
  & \targetimg{supp_cross_recon/target_Omniglot_0.pdf}
  & \targetimg{supp_cross_recon/target_Omniglot_1.pdf}
  & \targetimg{supp_cross_recon/target_Omniglot_2.pdf}
  \\
  \midrule
  \raisebox{8ex}{\rotatebox[origin=c]{90}{DooD}}
  & \reconimg{supp_cross_recon/dood-km_KMNIST_5.pdf}
  & \reconimg{supp_cross_recon/dood-km_KMNIST_6.pdf}
  & \reconimg{supp_cross_recon/dood-km_KMNIST_7.pdf}
  & \reconimg{supp_cross_recon/dood-km_KMNIST_8.pdf}
  & \reconimg{supp_cross_recon/dood-km_KMNIST_9.pdf}
  & \reconimg{supp_cross_recon/dood-km_KMNIST_13.pdf}
  & \reconimg{supp_cross_recon/dood-km_KMNIST_14.pdf}
  & \reconimg{supp_cross_recon/dood-km_KMNIST_15.pdf}
  & \reconimg{supp_cross_recon/dood-km_MNIST_0.pdf}
  & \reconimg{supp_cross_recon/dood-km_MNIST_1.pdf}
  & \reconimg{supp_cross_recon/dood-km_MNIST_2.pdf}
  & \reconimg{supp_cross_recon/dood-km_MNIST_3.pdf}
  & \reconimg{supp_cross_recon/dood-km_MNIST_4.pdf}
  & \reconimg{supp_cross_recon/dood-km_MNIST_5.pdf}
  & \reconimg{supp_cross_recon/dood-km_MNIST_6.pdf}
  & \reconimg{supp_cross_recon/dood-km_MNIST_7.pdf}
  & \reconimg{supp_cross_recon/dood-km_EMNIST_5.pdf}
  & \reconimg{supp_cross_recon/dood-km_EMNIST_6.pdf}
  & \reconimg{supp_cross_recon/dood-km_EMNIST_7.pdf}
  & \reconimg{supp_cross_recon/dood-km_EMNIST_8.pdf}
  & \reconimg{supp_cross_recon/dood-km_EMNIST_9.pdf}
  & \reconimg{supp_cross_recon/dood-km_EMNIST_10.pdf}
  & \reconimg{supp_cross_recon/dood-km_EMNIST_11.pdf}
  & \reconimg{supp_cross_recon/dood-km_EMNIST_12.pdf}
  & \reconimg{supp_cross_recon/dood-km_Quickdraw_5.pdf}
  & \reconimg{supp_cross_recon/dood-km_Quickdraw_6.pdf}
  & \reconimg{supp_cross_recon/dood-km_Quickdraw_7.pdf}
  & \reconimg{supp_cross_recon/dood-km_Quickdraw_8.pdf}
  & \reconimg{supp_cross_recon/dood-km_Quickdraw_9.pdf}
  & \reconimg{supp_cross_recon/dood-km_Quickdraw_10.pdf}
  & \reconimg{supp_cross_recon/dood-km_Quickdraw_11.pdf}
  & \reconimg{supp_cross_recon/dood-km_Quickdraw_12.pdf}
  & \reconimg{supp_cross_recon/dood-km_Omniglot_5.pdf}
  & \reconimg{supp_cross_recon/dood-km_Omniglot_6.pdf}
  & \reconimg{supp_cross_recon/dood-km_Omniglot_7.pdf}
  & \reconimg{supp_cross_recon/dood-km_Omniglot_8.pdf}
  & \reconimg{supp_cross_recon/dood-km_Omniglot_9.pdf}
  & \reconimg{supp_cross_recon/dood-km_Omniglot_0.pdf}
  & \reconimg{supp_cross_recon/dood-km_Omniglot_1.pdf}
  & \reconimg{supp_cross_recon/dood-km_Omniglot_2.pdf}
  \\
  \midrule
  \raisebox{8ex}{\rotatebox[origin=c]{90}{AIR}}
  & \reconimg{supp_cross_recon/dair-km_KMNIST_5.pdf}
  & \reconimg{supp_cross_recon/dair-km_KMNIST_6.pdf}
  & \reconimg{supp_cross_recon/dair-km_KMNIST_7.pdf}
  & \reconimg{supp_cross_recon/dair-km_KMNIST_8.pdf}
  & \reconimg{supp_cross_recon/dair-km_KMNIST_9.pdf}
  & \reconimg{supp_cross_recon/dair-km_KMNIST_13.pdf}
  & \reconimg{supp_cross_recon/dair-km_KMNIST_14.pdf}
  & \reconimg{supp_cross_recon/dair-km_KMNIST_15.pdf}
  & \reconimg{supp_cross_recon/dair-km_MNIST_0.pdf}
  & \reconimg{supp_cross_recon/dair-km_MNIST_1.pdf}
  & \reconimg{supp_cross_recon/dair-km_MNIST_2.pdf}
  & \reconimg{supp_cross_recon/dair-km_MNIST_3.pdf}
  & \reconimg{supp_cross_recon/dair-km_MNIST_4.pdf}
  & \reconimg{supp_cross_recon/dair-km_MNIST_5.pdf}
  & \reconimg{supp_cross_recon/dair-km_MNIST_6.pdf}
  & \reconimg{supp_cross_recon/dair-km_MNIST_7.pdf}
  & \reconimg{supp_cross_recon/dair-km_EMNIST_5.pdf}
  & \reconimg{supp_cross_recon/dair-km_EMNIST_6.pdf}
  & \reconimg{supp_cross_recon/dair-km_EMNIST_7.pdf}
  & \reconimg{supp_cross_recon/dair-km_EMNIST_8.pdf}
  & \reconimg{supp_cross_recon/dair-km_EMNIST_9.pdf}
  & \reconimg{supp_cross_recon/dair-km_EMNIST_10.pdf}
  & \reconimg{supp_cross_recon/dair-km_EMNIST_11.pdf}
  & \reconimg{supp_cross_recon/dair-km_EMNIST_12.pdf}
  & \reconimg{supp_cross_recon/dair-km_Quickdraw_5.pdf}
  & \reconimg{supp_cross_recon/dair-km_Quickdraw_6.pdf}
  & \reconimg{supp_cross_recon/dair-km_Quickdraw_7.pdf}
  & \reconimg{supp_cross_recon/dair-km_Quickdraw_8.pdf}
  & \reconimg{supp_cross_recon/dair-km_Quickdraw_9.pdf}
  & \reconimg{supp_cross_recon/dair-km_Quickdraw_10.pdf}
  & \reconimg{supp_cross_recon/dair-km_Quickdraw_11.pdf}
  & \reconimg{supp_cross_recon/dair-km_Quickdraw_12.pdf}
  & \reconimg{supp_cross_recon/dair-km_Omniglot_5.pdf}
  & \reconimg{supp_cross_recon/dair-km_Omniglot_6.pdf}
  & \reconimg{supp_cross_recon/dair-km_Omniglot_7.pdf}
  & \reconimg{supp_cross_recon/dair-km_Omniglot_8.pdf}
  & \reconimg{supp_cross_recon/dair-km_Omniglot_9.pdf}
  & \reconimg{supp_cross_recon/dair-km_Omniglot_0.pdf}
  & \reconimg{supp_cross_recon/dair-km_Omniglot_1.pdf}
  & \reconimg{supp_cross_recon/dair-km_Omniglot_2.pdf}
  \\
  \midrule
  \raisebox{8ex}{\rotatebox[origin=c]{90}{DooD-EG}}
  & \reconimg{supp_cross_recon/dood_eg-km_KMNIST_5.pdf}
  & \reconimg{supp_cross_recon/dood_eg-km_KMNIST_6.pdf}
  & \reconimg{supp_cross_recon/dood_eg-km_KMNIST_7.pdf}
  & \reconimg{supp_cross_recon/dood_eg-km_KMNIST_8.pdf}
  & \reconimg{supp_cross_recon/dood_eg-km_KMNIST_9.pdf}
  & \reconimg{supp_cross_recon/dood_eg-km_KMNIST_13.pdf}
  & \reconimg{supp_cross_recon/dood_eg-km_KMNIST_14.pdf}
  & \reconimg{supp_cross_recon/dood_eg-km_KMNIST_15.pdf}
  & \reconimg{supp_cross_recon/dood_eg-km_MNIST_0.pdf}
  & \reconimg{supp_cross_recon/dood_eg-km_MNIST_1.pdf}
  & \reconimg{supp_cross_recon/dood_eg-km_MNIST_2.pdf}
  & \reconimg{supp_cross_recon/dood_eg-km_MNIST_3.pdf}
  & \reconimg{supp_cross_recon/dood_eg-km_MNIST_4.pdf}
  & \reconimg{supp_cross_recon/dood_eg-km_MNIST_5.pdf}
  & \reconimg{supp_cross_recon/dood_eg-km_MNIST_6.pdf}
  & \reconimg{supp_cross_recon/dood_eg-km_MNIST_7.pdf}
  & \reconimg{supp_cross_recon/dood_eg-km_EMNIST_5.pdf}
  & \reconimg{supp_cross_recon/dood_eg-km_EMNIST_6.pdf}
  & \reconimg{supp_cross_recon/dood_eg-km_EMNIST_7.pdf}
  & \reconimg{supp_cross_recon/dood_eg-km_EMNIST_8.pdf}
  & \reconimg{supp_cross_recon/dood_eg-km_EMNIST_9.pdf}
  & \reconimg{supp_cross_recon/dood_eg-km_EMNIST_10.pdf}
  & \reconimg{supp_cross_recon/dood_eg-km_EMNIST_11.pdf}
  & \reconimg{supp_cross_recon/dood_eg-km_EMNIST_12.pdf}
  & \reconimg{supp_cross_recon/dood_eg-km_Quickdraw_5.pdf}
  & \reconimg{supp_cross_recon/dood_eg-km_Quickdraw_6.pdf}
  & \reconimg{supp_cross_recon/dood_eg-km_Quickdraw_7.pdf}
  & \reconimg{supp_cross_recon/dood_eg-km_Quickdraw_8.pdf}
  & \reconimg{supp_cross_recon/dood_eg-km_Quickdraw_9.pdf}
  & \reconimg{supp_cross_recon/dood_eg-km_Quickdraw_10.pdf}
  & \reconimg{supp_cross_recon/dood_eg-km_Quickdraw_11.pdf}
  & \reconimg{supp_cross_recon/dood_eg-km_Quickdraw_12.pdf}
  & \reconimg{supp_cross_recon/dood_eg-km_Omniglot_5.pdf}
  & \reconimg{supp_cross_recon/dood_eg-km_Omniglot_6.pdf}
  & \reconimg{supp_cross_recon/dood_eg-km_Omniglot_7.pdf}
  & \reconimg{supp_cross_recon/dood_eg-km_Omniglot_8.pdf}
  & \reconimg{supp_cross_recon/dood_eg-km_Omniglot_9.pdf}
  & \reconimg{supp_cross_recon/dood_eg-km_Omniglot_0.pdf}
  & \reconimg{supp_cross_recon/dood_eg-km_Omniglot_1.pdf}
  & \reconimg{supp_cross_recon/dood_eg-km_Omniglot_2.pdf}
  \\
\end{tabular}

}
    \captionof{figure}{KMNIST-trained model generalize to other datasets.}
    \label{fig:kmnist_generalization}
    \end{minipage}
  \end{sideways}
\end{center}

\newpage
\subsubsection{MNIST-trained models}

\begin{center}
  \begin{sideways}
    \begin{minipage}[c]{0.96\textheight}
    \scalebox{0.7}{\def\targetimg#1{\includegraphics[width=0.03\textwidth,trim={0 {0.6\textheight} 0 0},clip]{#1}}
\def\reconimg#1{\includegraphics[width=0.03\textwidth,trim={0 0 0 {0.097\textheight}},clip]{#1}}
\setlength\tabcolsep{1pt}
\begin{tabular}{@{}c@{\quad}*{8}{c}@{\hspace{20pt}}*{4}{*{8}{c}@{\hspace{10pt}}}@{\hspace{-10pt}}}
  (a)
  & \multicolumn{8}{c}{\quad MNIST \quad\(\bm{\to}\)\,}
  & \multicolumn{8}{c}{EMNIST}
  & \multicolumn{8}{c}{KMNIST}
  & \multicolumn{8}{c}{QuickDraw}
  & \multicolumn{8}{c}{Omniglot} \\
  \midrule
  $x$
  & \targetimg{supp_cross_recon/target_MNIST_0.pdf}
  & \targetimg{supp_cross_recon/target_MNIST_1.pdf}
  & \targetimg{supp_cross_recon/target_MNIST_2.pdf}
  & \targetimg{supp_cross_recon/target_MNIST_3.pdf}
  & \targetimg{supp_cross_recon/target_MNIST_4.pdf}
  & \targetimg{supp_cross_recon/target_MNIST_5.pdf}
  & \targetimg{supp_cross_recon/target_MNIST_6.pdf}
  & \targetimg{supp_cross_recon/target_MNIST_7.pdf}
  & \targetimg{supp_cross_recon/target_EMNIST_5.pdf}
  & \targetimg{supp_cross_recon/target_EMNIST_6.pdf}
  & \targetimg{supp_cross_recon/target_EMNIST_7.pdf}
  & \targetimg{supp_cross_recon/target_EMNIST_8.pdf}
  & \targetimg{supp_cross_recon/target_EMNIST_9.pdf}
  & \targetimg{supp_cross_recon/target_EMNIST_10.pdf}
  & \targetimg{supp_cross_recon/target_EMNIST_11.pdf}
  & \targetimg{supp_cross_recon/target_EMNIST_12.pdf}
  & \targetimg{supp_cross_recon/target_KMNIST_5.pdf}
  & \targetimg{supp_cross_recon/target_KMNIST_6.pdf}
  & \targetimg{supp_cross_recon/target_KMNIST_7.pdf}
  & \targetimg{supp_cross_recon/target_KMNIST_8.pdf}
  & \targetimg{supp_cross_recon/target_KMNIST_9.pdf}
  & \targetimg{supp_cross_recon/target_KMNIST_13.pdf}
  & \targetimg{supp_cross_recon/target_KMNIST_14.pdf}
  & \targetimg{supp_cross_recon/target_KMNIST_15.pdf}
  & \targetimg{supp_cross_recon/target_Quickdraw_5.pdf}
  & \targetimg{supp_cross_recon/target_Quickdraw_6.pdf}
  & \targetimg{supp_cross_recon/target_Quickdraw_7.pdf}
  & \targetimg{supp_cross_recon/target_Quickdraw_8.pdf}
  & \targetimg{supp_cross_recon/target_Quickdraw_9.pdf}
  & \targetimg{supp_cross_recon/target_Quickdraw_10.pdf}
  & \targetimg{supp_cross_recon/target_Quickdraw_11.pdf}
  & \targetimg{supp_cross_recon/target_Quickdraw_12.pdf}
  & \targetimg{supp_cross_recon/target_Omniglot_5.pdf}
  & \targetimg{supp_cross_recon/target_Omniglot_6.pdf}
  & \targetimg{supp_cross_recon/target_Omniglot_7.pdf}
  & \targetimg{supp_cross_recon/target_Omniglot_8.pdf}
  & \targetimg{supp_cross_recon/target_Omniglot_9.pdf}
  & \targetimg{supp_cross_recon/target_Omniglot_0.pdf}
  & \targetimg{supp_cross_recon/target_Omniglot_1.pdf}
  & \targetimg{supp_cross_recon/target_Omniglot_2.pdf}
  \\
  \midrule
  \raisebox{8ex}{\rotatebox[origin=c]{90}{DooD}}
  & \reconimg{supp_cross_recon/dood-mn_MNIST_0.pdf}
  & \reconimg{supp_cross_recon/dood-mn_MNIST_1.pdf}
  & \reconimg{supp_cross_recon/dood-mn_MNIST_2.pdf}
  & \reconimg{supp_cross_recon/dood-mn_MNIST_3.pdf}
  & \reconimg{supp_cross_recon/dood-mn_MNIST_4.pdf}
  & \reconimg{supp_cross_recon/dood-mn_MNIST_5.pdf}
  & \reconimg{supp_cross_recon/dood-mn_MNIST_6.pdf}
  & \reconimg{supp_cross_recon/dood-mn_MNIST_7.pdf}
  & \reconimg{supp_cross_recon/dood-mn_EMNIST_5.pdf}
  & \reconimg{supp_cross_recon/dood-mn_EMNIST_6.pdf}
  & \reconimg{supp_cross_recon/dood-mn_EMNIST_7.pdf}
  & \reconimg{supp_cross_recon/dood-mn_EMNIST_8.pdf}
  & \reconimg{supp_cross_recon/dood-mn_EMNIST_9.pdf}
  & \reconimg{supp_cross_recon/dood-mn_EMNIST_10.pdf}
  & \reconimg{supp_cross_recon/dood-mn_EMNIST_11.pdf}
  & \reconimg{supp_cross_recon/dood-mn_EMNIST_12.pdf}
  & \reconimg{supp_cross_recon/dood-mn_KMNIST_5.pdf}
  & \reconimg{supp_cross_recon/dood-mn_KMNIST_6.pdf}
  & \reconimg{supp_cross_recon/dood-mn_KMNIST_7.pdf}
  & \reconimg{supp_cross_recon/dood-mn_KMNIST_8.pdf}
  & \reconimg{supp_cross_recon/dood-mn_KMNIST_9.pdf}
  & \reconimg{supp_cross_recon/dood-mn_KMNIST_13.pdf}
  & \reconimg{supp_cross_recon/dood-mn_KMNIST_14.pdf}
  & \reconimg{supp_cross_recon/dood-mn_KMNIST_15.pdf}
  & \reconimg{supp_cross_recon/dood-mn_Quickdraw_5.pdf}
  & \reconimg{supp_cross_recon/dood-mn_Quickdraw_6.pdf}
  & \reconimg{supp_cross_recon/dood-mn_Quickdraw_7.pdf}
  & \reconimg{supp_cross_recon/dood-mn_Quickdraw_8.pdf}
  & \reconimg{supp_cross_recon/dood-mn_Quickdraw_9.pdf}
  & \reconimg{supp_cross_recon/dood-mn_Quickdraw_10.pdf}
  & \reconimg{supp_cross_recon/dood-mn_Quickdraw_11.pdf}
  & \reconimg{supp_cross_recon/dood-mn_Quickdraw_12.pdf}
  & \reconimg{supp_cross_recon/dood-mn_Omniglot_5.pdf}
  & \reconimg{supp_cross_recon/dood-mn_Omniglot_6.pdf}
  & \reconimg{supp_cross_recon/dood-mn_Omniglot_7.pdf}
  & \reconimg{supp_cross_recon/dood-mn_Omniglot_8.pdf}
  & \reconimg{supp_cross_recon/dood-mn_Omniglot_9.pdf}
  & \reconimg{supp_cross_recon/dood-mn_Omniglot_0.pdf}
  & \reconimg{supp_cross_recon/dood-mn_Omniglot_1.pdf}
  & \reconimg{supp_cross_recon/dood-mn_Omniglot_2.pdf}
  \\
  \midrule
  \raisebox{8ex}{\rotatebox[origin=c]{90}{AIR}}
  & \reconimg{supp_cross_recon/dair-mn_MNIST_0.pdf}
  & \reconimg{supp_cross_recon/dair-mn_MNIST_1.pdf}
  & \reconimg{supp_cross_recon/dair-mn_MNIST_2.pdf}
  & \reconimg{supp_cross_recon/dair-mn_MNIST_3.pdf}
  & \reconimg{supp_cross_recon/dair-mn_MNIST_4.pdf}
  & \reconimg{supp_cross_recon/dair-mn_MNIST_5.pdf}
  & \reconimg{supp_cross_recon/dair-mn_MNIST_6.pdf}
  & \reconimg{supp_cross_recon/dair-mn_MNIST_7.pdf}
  & \reconimg{supp_cross_recon/dair-mn_EMNIST_5.pdf}
  & \reconimg{supp_cross_recon/dair-mn_EMNIST_6.pdf}
  & \reconimg{supp_cross_recon/dair-mn_EMNIST_7.pdf}
  & \reconimg{supp_cross_recon/dair-mn_EMNIST_8.pdf}
  & \reconimg{supp_cross_recon/dair-mn_EMNIST_9.pdf}
  & \reconimg{supp_cross_recon/dair-mn_EMNIST_10.pdf}
  & \reconimg{supp_cross_recon/dair-mn_EMNIST_11.pdf}
  & \reconimg{supp_cross_recon/dair-mn_EMNIST_12.pdf}
  & \reconimg{supp_cross_recon/dair-mn_KMNIST_5.pdf}
  & \reconimg{supp_cross_recon/dair-mn_KMNIST_6.pdf}
  & \reconimg{supp_cross_recon/dair-mn_KMNIST_7.pdf}
  & \reconimg{supp_cross_recon/dair-mn_KMNIST_8.pdf}
  & \reconimg{supp_cross_recon/dair-mn_KMNIST_9.pdf}
  & \reconimg{supp_cross_recon/dair-mn_KMNIST_13.pdf}
  & \reconimg{supp_cross_recon/dair-mn_KMNIST_14.pdf}
  & \reconimg{supp_cross_recon/dair-mn_KMNIST_15.pdf}
  & \reconimg{supp_cross_recon/dair-mn_Quickdraw_5.pdf}
  & \reconimg{supp_cross_recon/dair-mn_Quickdraw_6.pdf}
  & \reconimg{supp_cross_recon/dair-mn_Quickdraw_7.pdf}
  & \reconimg{supp_cross_recon/dair-mn_Quickdraw_8.pdf}
  & \reconimg{supp_cross_recon/dair-mn_Quickdraw_9.pdf}
  & \reconimg{supp_cross_recon/dair-mn_Quickdraw_10.pdf}
  & \reconimg{supp_cross_recon/dair-mn_Quickdraw_11.pdf}
  & \reconimg{supp_cross_recon/dair-mn_Quickdraw_12.pdf}
  & \reconimg{supp_cross_recon/dair-mn_Omniglot_5.pdf}
  & \reconimg{supp_cross_recon/dair-mn_Omniglot_6.pdf}
  & \reconimg{supp_cross_recon/dair-mn_Omniglot_7.pdf}
  & \reconimg{supp_cross_recon/dair-mn_Omniglot_8.pdf}
  & \reconimg{supp_cross_recon/dair-mn_Omniglot_9.pdf}
  & \reconimg{supp_cross_recon/dair-mn_Omniglot_0.pdf}
  & \reconimg{supp_cross_recon/dair-mn_Omniglot_1.pdf}
  & \reconimg{supp_cross_recon/dair-mn_Omniglot_2.pdf}
  \\
  \midrule
  \raisebox{8ex}{\rotatebox[origin=c]{90}{DooD-EG}}
  & \reconimg{supp_cross_recon/dood_eg-mn_MNIST_0.pdf}
  & \reconimg{supp_cross_recon/dood_eg-mn_MNIST_1.pdf}
  & \reconimg{supp_cross_recon/dood_eg-mn_MNIST_2.pdf}
  & \reconimg{supp_cross_recon/dood_eg-mn_MNIST_3.pdf}
  & \reconimg{supp_cross_recon/dood_eg-mn_MNIST_4.pdf}
  & \reconimg{supp_cross_recon/dood_eg-mn_MNIST_5.pdf}
  & \reconimg{supp_cross_recon/dood_eg-mn_MNIST_6.pdf}
  & \reconimg{supp_cross_recon/dood_eg-mn_MNIST_7.pdf}
  & \reconimg{supp_cross_recon/dood_eg-mn_EMNIST_5.pdf}
  & \reconimg{supp_cross_recon/dood_eg-mn_EMNIST_6.pdf}
  & \reconimg{supp_cross_recon/dood_eg-mn_EMNIST_7.pdf}
  & \reconimg{supp_cross_recon/dood_eg-mn_EMNIST_8.pdf}
  & \reconimg{supp_cross_recon/dood_eg-mn_EMNIST_9.pdf}
  & \reconimg{supp_cross_recon/dood_eg-mn_EMNIST_10.pdf}
  & \reconimg{supp_cross_recon/dood_eg-mn_EMNIST_11.pdf}
  & \reconimg{supp_cross_recon/dood_eg-mn_EMNIST_12.pdf}
  & \reconimg{supp_cross_recon/dood_eg-mn_KMNIST_5.pdf}
  & \reconimg{supp_cross_recon/dood_eg-mn_KMNIST_6.pdf}
  & \reconimg{supp_cross_recon/dood_eg-mn_KMNIST_7.pdf}
  & \reconimg{supp_cross_recon/dood_eg-mn_KMNIST_8.pdf}
  & \reconimg{supp_cross_recon/dood_eg-mn_KMNIST_9.pdf}
  & \reconimg{supp_cross_recon/dood_eg-mn_KMNIST_13.pdf}
  & \reconimg{supp_cross_recon/dood_eg-mn_KMNIST_14.pdf}
  & \reconimg{supp_cross_recon/dood_eg-mn_KMNIST_15.pdf}
  & \reconimg{supp_cross_recon/dood_eg-mn_Quickdraw_5.pdf}
  & \reconimg{supp_cross_recon/dood_eg-mn_Quickdraw_6.pdf}
  & \reconimg{supp_cross_recon/dood_eg-mn_Quickdraw_7.pdf}
  & \reconimg{supp_cross_recon/dood_eg-mn_Quickdraw_8.pdf}
  & \reconimg{supp_cross_recon/dood_eg-mn_Quickdraw_9.pdf}
  & \reconimg{supp_cross_recon/dood_eg-mn_Quickdraw_10.pdf}
  & \reconimg{supp_cross_recon/dood_eg-mn_Quickdraw_11.pdf}
  & \reconimg{supp_cross_recon/dood_eg-mn_Quickdraw_12.pdf}
  & \reconimg{supp_cross_recon/dood_eg-mn_Omniglot_5.pdf}
  & \reconimg{supp_cross_recon/dood_eg-mn_Omniglot_6.pdf}
  & \reconimg{supp_cross_recon/dood_eg-mn_Omniglot_7.pdf}
  & \reconimg{supp_cross_recon/dood_eg-mn_Omniglot_8.pdf}
  & \reconimg{supp_cross_recon/dood_eg-mn_Omniglot_9.pdf}
  & \reconimg{supp_cross_recon/dood_eg-mn_Omniglot_0.pdf}
  & \reconimg{supp_cross_recon/dood_eg-mn_Omniglot_1.pdf}
  & \reconimg{supp_cross_recon/dood_eg-mn_Omniglot_2.pdf}
  \\
\end{tabular}

}
    \captionof{figure}{MNIST-trained model generalize to other datasets.}
    \label{fig:mnist_generalization_xl}
    \end{minipage}
  \end{sideways}
\end{center}

\newpage
\subsubsection{Omniglot-trained models}
\begin{center}
  \begin{sideways}
    \begin{minipage}[c]{0.96\textheight}
    \scalebox{0.7}{\def\targetimg#1{\includegraphics[width=0.03\textwidth,trim={0 {0.6\textheight} 0 0},clip]{#1}}
\def\reconimg#1{\includegraphics[width=0.03\textwidth,trim={0 0 0 {0.097\textheight}},clip]{#1}}
\setlength\tabcolsep{1pt}
\begin{tabular}{@{}c@{\quad}*{8}{c}@{\hspace{20pt}}*{4}{*{8}{c}@{\hspace{10pt}}}@{\hspace{-10pt}}}
  (a)
  & \multicolumn{8}{c}{\quad Omniglot \quad\(\bm{\to}\)\,}
  & \multicolumn{8}{c}{MNIST}
  & \multicolumn{8}{c}{EMNIST}
  & \multicolumn{8}{c}{KMNIST}
  & \multicolumn{8}{c}{QuickDraw} \\
  \midrule
  $x$
  & \targetimg{supp_cross_recon/target_Omniglot_5.pdf}
  & \targetimg{supp_cross_recon/target_Omniglot_6.pdf}
  & \targetimg{supp_cross_recon/target_Omniglot_7.pdf}
  & \targetimg{supp_cross_recon/target_Omniglot_8.pdf}
  & \targetimg{supp_cross_recon/target_Omniglot_9.pdf}
  & \targetimg{supp_cross_recon/target_Omniglot_0.pdf}
  & \targetimg{supp_cross_recon/target_Omniglot_1.pdf}
  & \targetimg{supp_cross_recon/target_Omniglot_2.pdf}
  & \targetimg{supp_cross_recon/target_MNIST_0.pdf}
  & \targetimg{supp_cross_recon/target_MNIST_1.pdf}
  & \targetimg{supp_cross_recon/target_MNIST_2.pdf}
  & \targetimg{supp_cross_recon/target_MNIST_3.pdf}
  & \targetimg{supp_cross_recon/target_MNIST_4.pdf}
  & \targetimg{supp_cross_recon/target_MNIST_5.pdf}
  & \targetimg{supp_cross_recon/target_MNIST_6.pdf}
  & \targetimg{supp_cross_recon/target_MNIST_7.pdf}
  & \targetimg{supp_cross_recon/target_EMNIST_5.pdf}
  & \targetimg{supp_cross_recon/target_EMNIST_6.pdf}
  & \targetimg{supp_cross_recon/target_EMNIST_7.pdf}
  & \targetimg{supp_cross_recon/target_EMNIST_8.pdf}
  & \targetimg{supp_cross_recon/target_EMNIST_9.pdf}
  & \targetimg{supp_cross_recon/target_EMNIST_10.pdf}
  & \targetimg{supp_cross_recon/target_EMNIST_11.pdf}
  & \targetimg{supp_cross_recon/target_EMNIST_12.pdf}
  & \targetimg{supp_cross_recon/target_KMNIST_5.pdf}
  & \targetimg{supp_cross_recon/target_KMNIST_6.pdf}
  & \targetimg{supp_cross_recon/target_KMNIST_7.pdf}
  & \targetimg{supp_cross_recon/target_KMNIST_8.pdf}
  & \targetimg{supp_cross_recon/target_KMNIST_9.pdf}
  & \targetimg{supp_cross_recon/target_KMNIST_13.pdf}
  & \targetimg{supp_cross_recon/target_KMNIST_14.pdf}
  & \targetimg{supp_cross_recon/target_KMNIST_15.pdf}
  & \targetimg{supp_cross_recon/target_Quickdraw_5.pdf}
  & \targetimg{supp_cross_recon/target_Quickdraw_6.pdf}
  & \targetimg{supp_cross_recon/target_Quickdraw_7.pdf}
  & \targetimg{supp_cross_recon/target_Quickdraw_8.pdf}
  & \targetimg{supp_cross_recon/target_Quickdraw_9.pdf}
  & \targetimg{supp_cross_recon/target_Quickdraw_10.pdf}
  & \targetimg{supp_cross_recon/target_Quickdraw_11.pdf}
  & \targetimg{supp_cross_recon/target_Quickdraw_12.pdf}
  \\
  \midrule
  \raisebox{8ex}{\rotatebox[origin=c]{90}{DooD}}
  & \reconimg{supp_cross_recon/dood-om_Omniglot_5.pdf}
  & \reconimg{supp_cross_recon/dood-om_Omniglot_6.pdf}
  & \reconimg{supp_cross_recon/dood-om_Omniglot_7.pdf}
  & \reconimg{supp_cross_recon/dood-om_Omniglot_8.pdf}
  & \reconimg{supp_cross_recon/dood-om_Omniglot_9.pdf}
  & \reconimg{supp_cross_recon/dood-om_Omniglot_0.pdf}
  & \reconimg{supp_cross_recon/dood-om_Omniglot_1.pdf}
  & \reconimg{supp_cross_recon/dood-om_Omniglot_2.pdf}
  & \reconimg{supp_cross_recon/dood-om_MNIST_0.pdf}
  & \reconimg{supp_cross_recon/dood-om_MNIST_1.pdf}
  & \reconimg{supp_cross_recon/dood-om_MNIST_2.pdf}
  & \reconimg{supp_cross_recon/dood-om_MNIST_3.pdf}
  & \reconimg{supp_cross_recon/dood-om_MNIST_4.pdf}
  & \reconimg{supp_cross_recon/dood-om_MNIST_5.pdf}
  & \reconimg{supp_cross_recon/dood-om_MNIST_6.pdf}
  & \reconimg{supp_cross_recon/dood-om_MNIST_7.pdf}
  & \reconimg{supp_cross_recon/dood-om_EMNIST_5.pdf}
  & \reconimg{supp_cross_recon/dood-om_EMNIST_6.pdf}
  & \reconimg{supp_cross_recon/dood-om_EMNIST_7.pdf}
  & \reconimg{supp_cross_recon/dood-om_EMNIST_8.pdf}
  & \reconimg{supp_cross_recon/dood-om_EMNIST_9.pdf}
  & \reconimg{supp_cross_recon/dood-om_EMNIST_10.pdf}
  & \reconimg{supp_cross_recon/dood-om_EMNIST_11.pdf}
  & \reconimg{supp_cross_recon/dood-om_EMNIST_12.pdf}
  & \reconimg{supp_cross_recon/dood-om_KMNIST_5.pdf}
  & \reconimg{supp_cross_recon/dood-om_KMNIST_6.pdf}
  & \reconimg{supp_cross_recon/dood-om_KMNIST_7.pdf}
  & \reconimg{supp_cross_recon/dood-om_KMNIST_8.pdf}
  & \reconimg{supp_cross_recon/dood-om_KMNIST_9.pdf}
  & \reconimg{supp_cross_recon/dood-om_KMNIST_13.pdf}
  & \reconimg{supp_cross_recon/dood-om_KMNIST_14.pdf}
  & \reconimg{supp_cross_recon/dood-om_KMNIST_15.pdf}
  & \reconimg{supp_cross_recon/dood-om_Quickdraw_5.pdf}
  & \reconimg{supp_cross_recon/dood-om_Quickdraw_6.pdf}
  & \reconimg{supp_cross_recon/dood-om_Quickdraw_7.pdf}
  & \reconimg{supp_cross_recon/dood-om_Quickdraw_8.pdf}
  & \reconimg{supp_cross_recon/dood-om_Quickdraw_9.pdf}
  & \reconimg{supp_cross_recon/dood-om_Quickdraw_10.pdf}
  & \reconimg{supp_cross_recon/dood-om_Quickdraw_11.pdf}
  & \reconimg{supp_cross_recon/dood-om_Quickdraw_12.pdf}
  \\
  \midrule
  \raisebox{8ex}{\rotatebox[origin=c]{90}{AIR}}
  & \reconimg{supp_cross_recon/dair-om_Omniglot_5.pdf}
  & \reconimg{supp_cross_recon/dair-om_Omniglot_6.pdf}
  & \reconimg{supp_cross_recon/dair-om_Omniglot_7.pdf}
  & \reconimg{supp_cross_recon/dair-om_Omniglot_8.pdf}
  & \reconimg{supp_cross_recon/dair-om_Omniglot_9.pdf}
  & \reconimg{supp_cross_recon/dair-om_Omniglot_0.pdf}
  & \reconimg{supp_cross_recon/dair-om_Omniglot_1.pdf}
  & \reconimg{supp_cross_recon/dair-om_Omniglot_2.pdf}
  & \reconimg{supp_cross_recon/dair-om_MNIST_0.pdf}
  & \reconimg{supp_cross_recon/dair-om_MNIST_1.pdf}
  & \reconimg{supp_cross_recon/dair-om_MNIST_2.pdf}
  & \reconimg{supp_cross_recon/dair-om_MNIST_3.pdf}
  & \reconimg{supp_cross_recon/dair-om_MNIST_4.pdf}
  & \reconimg{supp_cross_recon/dair-om_MNIST_5.pdf}
  & \reconimg{supp_cross_recon/dair-om_MNIST_6.pdf}
  & \reconimg{supp_cross_recon/dair-om_MNIST_7.pdf}
  & \reconimg{supp_cross_recon/dair-om_EMNIST_5.pdf}
  & \reconimg{supp_cross_recon/dair-om_EMNIST_6.pdf}
  & \reconimg{supp_cross_recon/dair-om_EMNIST_7.pdf}
  & \reconimg{supp_cross_recon/dair-om_EMNIST_8.pdf}
  & \reconimg{supp_cross_recon/dair-om_EMNIST_9.pdf}
  & \reconimg{supp_cross_recon/dair-om_EMNIST_10.pdf}
  & \reconimg{supp_cross_recon/dair-om_EMNIST_11.pdf}
  & \reconimg{supp_cross_recon/dair-om_EMNIST_12.pdf}
  & \reconimg{supp_cross_recon/dair-om_KMNIST_5.pdf}
  & \reconimg{supp_cross_recon/dair-om_KMNIST_6.pdf}
  & \reconimg{supp_cross_recon/dair-om_KMNIST_7.pdf}
  & \reconimg{supp_cross_recon/dair-om_KMNIST_8.pdf}
  & \reconimg{supp_cross_recon/dair-om_KMNIST_9.pdf}
  & \reconimg{supp_cross_recon/dair-om_KMNIST_13.pdf}
  & \reconimg{supp_cross_recon/dair-om_KMNIST_14.pdf}
  & \reconimg{supp_cross_recon/dair-om_KMNIST_15.pdf}
  & \reconimg{supp_cross_recon/dair-om_Quickdraw_5.pdf}
  & \reconimg{supp_cross_recon/dair-om_Quickdraw_6.pdf}
  & \reconimg{supp_cross_recon/dair-om_Quickdraw_7.pdf}
  & \reconimg{supp_cross_recon/dair-om_Quickdraw_8.pdf}
  & \reconimg{supp_cross_recon/dair-om_Quickdraw_9.pdf}
  & \reconimg{supp_cross_recon/dair-om_Quickdraw_10.pdf}
  & \reconimg{supp_cross_recon/dair-om_Quickdraw_11.pdf}
  & \reconimg{supp_cross_recon/dair-om_Quickdraw_12.pdf}
  \\
\end{tabular}

}
    
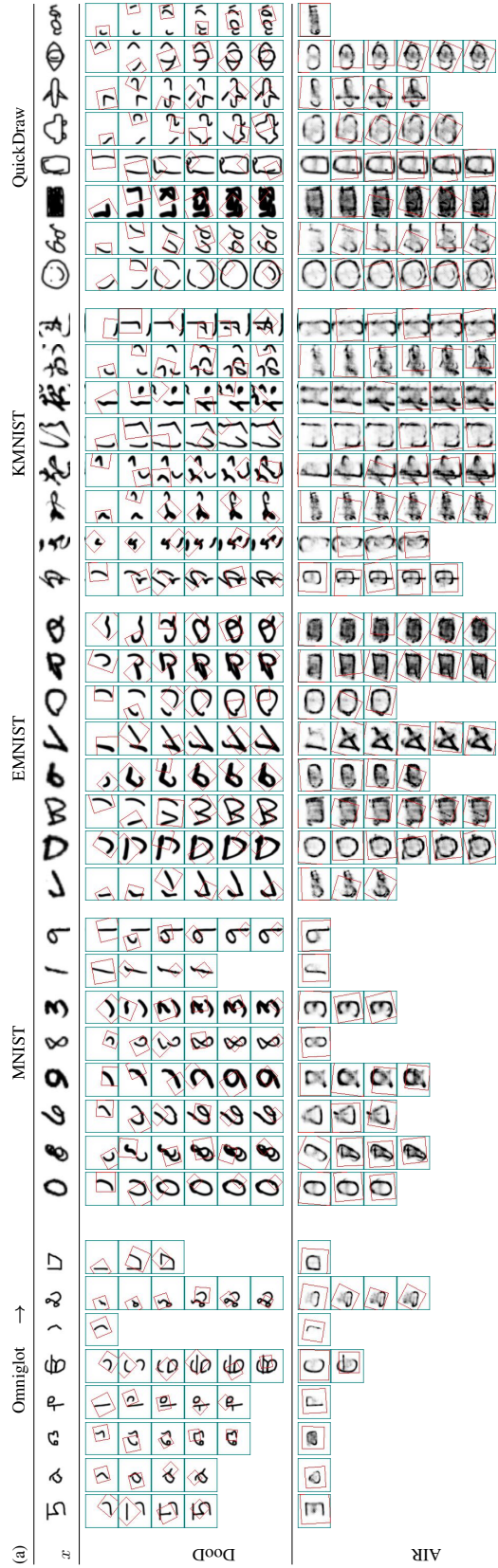
\captionof{figure}{Omniglot-trained model generalize to other datasets.}
    \label{fig:omniglot_generalization}
    \end{minipage}
  \end{sideways}
\end{center}

\newpage
\subsubsection{QuickDraw-trained models}
\begin{center}
  \begin{sideways}
    \begin{minipage}[c]{0.96\textheight}
    \scalebox{0.7}{\def\targetimg#1{\includegraphics[width=0.03\textwidth,trim={0 {0.6\textheight} 0 0},clip]{#1}}
\def\reconimg#1{\includegraphics[width=0.03\textwidth,trim={0 0 0 {0.097\textheight}},clip]{#1}}
\setlength\tabcolsep{1pt}
\begin{tabular}{@{}c@{\quad}*{8}{c}@{\hspace{20pt}}*{4}{*{8}{c}@{\hspace{10pt}}}@{\hspace{-10pt}}}
  (a)
  & \multicolumn{8}{c}{\quad QuickDraw \quad\(\bm{\to}\)\,}
  & \multicolumn{8}{c}{MNIST}
  & \multicolumn{8}{c}{EMNIST}
  & \multicolumn{8}{c}{KMNIST}
  & \multicolumn{8}{c}{Omniglot} \\
  \midrule
  $x$
  & \targetimg{supp_cross_recon/target_Quickdraw_5.pdf}
  & \targetimg{supp_cross_recon/target_Quickdraw_6.pdf}
  & \targetimg{supp_cross_recon/target_Quickdraw_7.pdf}
  & \targetimg{supp_cross_recon/target_Quickdraw_8.pdf}
  & \targetimg{supp_cross_recon/target_Quickdraw_9.pdf}
  & \targetimg{supp_cross_recon/target_Quickdraw_10.pdf}
  & \targetimg{supp_cross_recon/target_Quickdraw_11.pdf}
  & \targetimg{supp_cross_recon/target_Quickdraw_12.pdf}
  & \targetimg{supp_cross_recon/target_MNIST_0.pdf}
  & \targetimg{supp_cross_recon/target_MNIST_1.pdf}
  & \targetimg{supp_cross_recon/target_MNIST_2.pdf}
  & \targetimg{supp_cross_recon/target_MNIST_3.pdf}
  & \targetimg{supp_cross_recon/target_MNIST_4.pdf}
  & \targetimg{supp_cross_recon/target_MNIST_5.pdf}
  & \targetimg{supp_cross_recon/target_MNIST_6.pdf}
  & \targetimg{supp_cross_recon/target_MNIST_7.pdf}
  & \targetimg{supp_cross_recon/target_EMNIST_5.pdf}
  & \targetimg{supp_cross_recon/target_EMNIST_6.pdf}
  & \targetimg{supp_cross_recon/target_EMNIST_7.pdf}
  & \targetimg{supp_cross_recon/target_EMNIST_8.pdf}
  & \targetimg{supp_cross_recon/target_EMNIST_9.pdf}
  & \targetimg{supp_cross_recon/target_EMNIST_10.pdf}
  & \targetimg{supp_cross_recon/target_EMNIST_11.pdf}
  & \targetimg{supp_cross_recon/target_EMNIST_12.pdf}
  & \targetimg{supp_cross_recon/target_KMNIST_5.pdf}
  & \targetimg{supp_cross_recon/target_KMNIST_6.pdf}
  & \targetimg{supp_cross_recon/target_KMNIST_7.pdf}
  & \targetimg{supp_cross_recon/target_KMNIST_8.pdf}
  & \targetimg{supp_cross_recon/target_KMNIST_9.pdf}
  & \targetimg{supp_cross_recon/target_KMNIST_13.pdf}
  & \targetimg{supp_cross_recon/target_KMNIST_14.pdf}
  & \targetimg{supp_cross_recon/target_KMNIST_15.pdf}
  & \targetimg{supp_cross_recon/target_Omniglot_5.pdf}
  & \targetimg{supp_cross_recon/target_Omniglot_6.pdf}
  & \targetimg{supp_cross_recon/target_Omniglot_7.pdf}
  & \targetimg{supp_cross_recon/target_Omniglot_8.pdf}
  & \targetimg{supp_cross_recon/target_Omniglot_9.pdf}
  & \targetimg{supp_cross_recon/target_Omniglot_0.pdf}
  & \targetimg{supp_cross_recon/target_Omniglot_1.pdf}
  & \targetimg{supp_cross_recon/target_Omniglot_2.pdf}
  \\
  \midrule
  \raisebox{8ex}{\rotatebox[origin=c]{90}{DooD}}
  & \reconimg{supp_cross_recon/dood-qd_Quickdraw_5.pdf}
  & \reconimg{supp_cross_recon/dood-qd_Quickdraw_6.pdf}
  & \reconimg{supp_cross_recon/dood-qd_Quickdraw_7.pdf}
  & \reconimg{supp_cross_recon/dood-qd_Quickdraw_8.pdf}
  & \reconimg{supp_cross_recon/dood-qd_Quickdraw_9.pdf}
  & \reconimg{supp_cross_recon/dood-qd_Quickdraw_10.pdf}
  & \reconimg{supp_cross_recon/dood-qd_Quickdraw_11.pdf}
  & \reconimg{supp_cross_recon/dood-qd_Quickdraw_12.pdf}
  & \reconimg{supp_cross_recon/dood-qd_MNIST_0.pdf}
  & \reconimg{supp_cross_recon/dood-qd_MNIST_1.pdf}
  & \reconimg{supp_cross_recon/dood-qd_MNIST_2.pdf}
  & \reconimg{supp_cross_recon/dood-qd_MNIST_3.pdf}
  & \reconimg{supp_cross_recon/dood-qd_MNIST_4.pdf}
  & \reconimg{supp_cross_recon/dood-qd_MNIST_5.pdf}
  & \reconimg{supp_cross_recon/dood-qd_MNIST_6.pdf}
  & \reconimg{supp_cross_recon/dood-qd_MNIST_7.pdf}
  & \reconimg{supp_cross_recon/dood-qd_EMNIST_5.pdf}
  & \reconimg{supp_cross_recon/dood-qd_EMNIST_6.pdf}
  & \reconimg{supp_cross_recon/dood-qd_EMNIST_7.pdf}
  & \reconimg{supp_cross_recon/dood-qd_EMNIST_8.pdf}
  & \reconimg{supp_cross_recon/dood-qd_EMNIST_9.pdf}
  & \reconimg{supp_cross_recon/dood-qd_EMNIST_10.pdf}
  & \reconimg{supp_cross_recon/dood-qd_EMNIST_11.pdf}
  & \reconimg{supp_cross_recon/dood-qd_EMNIST_12.pdf}
  & \reconimg{supp_cross_recon/dood-qd_KMNIST_5.pdf}
  & \reconimg{supp_cross_recon/dood-qd_KMNIST_6.pdf}
  & \reconimg{supp_cross_recon/dood-qd_KMNIST_7.pdf}
  & \reconimg{supp_cross_recon/dood-qd_KMNIST_8.pdf}
  & \reconimg{supp_cross_recon/dood-qd_KMNIST_9.pdf}
  & \reconimg{supp_cross_recon/dood-qd_KMNIST_13.pdf}
  & \reconimg{supp_cross_recon/dood-qd_KMNIST_14.pdf}
  & \reconimg{supp_cross_recon/dood-qd_KMNIST_15.pdf}
  & \reconimg{supp_cross_recon/dood-qd_Omniglot_5.pdf}
  & \reconimg{supp_cross_recon/dood-qd_Omniglot_6.pdf}
  & \reconimg{supp_cross_recon/dood-qd_Omniglot_7.pdf}
  & \reconimg{supp_cross_recon/dood-qd_Omniglot_8.pdf}
  & \reconimg{supp_cross_recon/dood-qd_Omniglot_9.pdf}
  & \reconimg{supp_cross_recon/dood-qd_Omniglot_0.pdf}
  & \reconimg{supp_cross_recon/dood-qd_Omniglot_1.pdf}
  & \reconimg{supp_cross_recon/dood-qd_Omniglot_2.pdf}
  \\
  \midrule
  \raisebox{8ex}{\rotatebox[origin=c]{90}{AIR}}
  & \reconimg{supp_cross_recon/dair-qd_Quickdraw_5.pdf}
  & \reconimg{supp_cross_recon/dair-qd_Quickdraw_6.pdf}
  & \reconimg{supp_cross_recon/dair-qd_Quickdraw_7.pdf}
  & \reconimg{supp_cross_recon/dair-qd_Quickdraw_8.pdf}
  & \reconimg{supp_cross_recon/dair-qd_Quickdraw_9.pdf}
  & \reconimg{supp_cross_recon/dair-qd_Quickdraw_10.pdf}
  & \reconimg{supp_cross_recon/dair-qd_Quickdraw_11.pdf}
  & \reconimg{supp_cross_recon/dair-qd_Quickdraw_12.pdf}
  & \reconimg{supp_cross_recon/dair-qd_MNIST_0.pdf}
  & \reconimg{supp_cross_recon/dair-qd_MNIST_1.pdf}
  & \reconimg{supp_cross_recon/dair-qd_MNIST_2.pdf}
  & \reconimg{supp_cross_recon/dair-qd_MNIST_3.pdf}
  & \reconimg{supp_cross_recon/dair-qd_MNIST_4.pdf}
  & \reconimg{supp_cross_recon/dair-qd_MNIST_5.pdf}
  & \reconimg{supp_cross_recon/dair-qd_MNIST_6.pdf}
  & \reconimg{supp_cross_recon/dair-qd_MNIST_7.pdf}
  & \reconimg{supp_cross_recon/dair-qd_EMNIST_5.pdf}
  & \reconimg{supp_cross_recon/dair-qd_EMNIST_6.pdf}
  & \reconimg{supp_cross_recon/dair-qd_EMNIST_7.pdf}
  & \reconimg{supp_cross_recon/dair-qd_EMNIST_8.pdf}
  & \reconimg{supp_cross_recon/dair-qd_EMNIST_9.pdf}
  & \reconimg{supp_cross_recon/dair-qd_EMNIST_10.pdf}
  & \reconimg{supp_cross_recon/dair-qd_EMNIST_11.pdf}
  & \reconimg{supp_cross_recon/dair-qd_EMNIST_12.pdf}
  & \reconimg{supp_cross_recon/dair-qd_KMNIST_5.pdf}
  & \reconimg{supp_cross_recon/dair-qd_KMNIST_6.pdf}
  & \reconimg{supp_cross_recon/dair-qd_KMNIST_7.pdf}
  & \reconimg{supp_cross_recon/dair-qd_KMNIST_8.pdf}
  & \reconimg{supp_cross_recon/dair-qd_KMNIST_9.pdf}
  & \reconimg{supp_cross_recon/dair-qd_KMNIST_13.pdf}
  & \reconimg{supp_cross_recon/dair-qd_KMNIST_14.pdf}
  & \reconimg{supp_cross_recon/dair-qd_KMNIST_15.pdf}
  & \reconimg{supp_cross_recon/dair-qd_Omniglot_5.pdf}
  & \reconimg{supp_cross_recon/dair-qd_Omniglot_6.pdf}
  & \reconimg{supp_cross_recon/dair-qd_Omniglot_7.pdf}
  & \reconimg{supp_cross_recon/dair-qd_Omniglot_8.pdf}
  & \reconimg{supp_cross_recon/dair-qd_Omniglot_9.pdf}
  & \reconimg{supp_cross_recon/dair-qd_Omniglot_0.pdf}
  & \reconimg{supp_cross_recon/dair-qd_Omniglot_1.pdf}
  & \reconimg{supp_cross_recon/dair-qd_Omniglot_2.pdf}
  \\
  \midrule
  \raisebox{8ex}{\rotatebox[origin=c]{90}{DooD-EG}}
  & \reconimg{supp_cross_recon/dood_eg-qd_Quickdraw_5.pdf}
  & \reconimg{supp_cross_recon/dood_eg-qd_Quickdraw_6.pdf}
  & \reconimg{supp_cross_recon/dood_eg-qd_Quickdraw_7.pdf}
  & \reconimg{supp_cross_recon/dood_eg-qd_Quickdraw_8.pdf}
  & \reconimg{supp_cross_recon/dood_eg-qd_Quickdraw_9.pdf}
  & \reconimg{supp_cross_recon/dood_eg-qd_Quickdraw_10.pdf}
  & \reconimg{supp_cross_recon/dood_eg-qd_Quickdraw_11.pdf}
  & \reconimg{supp_cross_recon/dood_eg-qd_Quickdraw_12.pdf}
  & \reconimg{supp_cross_recon/dood_eg-qd_MNIST_0.pdf}
  & \reconimg{supp_cross_recon/dood_eg-qd_MNIST_1.pdf}
  & \reconimg{supp_cross_recon/dood_eg-qd_MNIST_2.pdf}
  & \reconimg{supp_cross_recon/dood_eg-qd_MNIST_3.pdf}
  & \reconimg{supp_cross_recon/dood_eg-qd_MNIST_4.pdf}
  & \reconimg{supp_cross_recon/dood_eg-qd_MNIST_5.pdf}
  & \reconimg{supp_cross_recon/dood_eg-qd_MNIST_6.pdf}
  & \reconimg{supp_cross_recon/dood_eg-qd_MNIST_7.pdf}
  & \reconimg{supp_cross_recon/dood_eg-qd_EMNIST_5.pdf}
  & \reconimg{supp_cross_recon/dood_eg-qd_EMNIST_6.pdf}
  & \reconimg{supp_cross_recon/dood_eg-qd_EMNIST_7.pdf}
  & \reconimg{supp_cross_recon/dood_eg-qd_EMNIST_8.pdf}
  & \reconimg{supp_cross_recon/dood_eg-qd_EMNIST_9.pdf}
  & \reconimg{supp_cross_recon/dood_eg-qd_EMNIST_10.pdf}
  & \reconimg{supp_cross_recon/dood_eg-qd_EMNIST_11.pdf}
  & \reconimg{supp_cross_recon/dood_eg-qd_EMNIST_12.pdf}
  & \reconimg{supp_cross_recon/dood_eg-qd_KMNIST_5.pdf}
  & \reconimg{supp_cross_recon/dood_eg-qd_KMNIST_6.pdf}
  & \reconimg{supp_cross_recon/dood_eg-qd_KMNIST_7.pdf}
  & \reconimg{supp_cross_recon/dood_eg-qd_KMNIST_8.pdf}
  & \reconimg{supp_cross_recon/dood_eg-qd_KMNIST_9.pdf}
  & \reconimg{supp_cross_recon/dood_eg-qd_KMNIST_13.pdf}
  & \reconimg{supp_cross_recon/dood_eg-qd_KMNIST_14.pdf}
  & \reconimg{supp_cross_recon/dood_eg-qd_KMNIST_15.pdf}
  & \reconimg{supp_cross_recon/dood_eg-qd_Omniglot_5.pdf}
  & \reconimg{supp_cross_recon/dood_eg-qd_Omniglot_6.pdf}
  & \reconimg{supp_cross_recon/dood_eg-qd_Omniglot_7.pdf}
  & \reconimg{supp_cross_recon/dood_eg-qd_Omniglot_8.pdf}
  & \reconimg{supp_cross_recon/dood_eg-qd_Omniglot_9.pdf}
  & \reconimg{supp_cross_recon/dood_eg-qd_Omniglot_0.pdf}
  & \reconimg{supp_cross_recon/dood_eg-qd_Omniglot_1.pdf}
  & \reconimg{supp_cross_recon/dood_eg-qd_Omniglot_2.pdf}
  \\
\end{tabular}

}
    
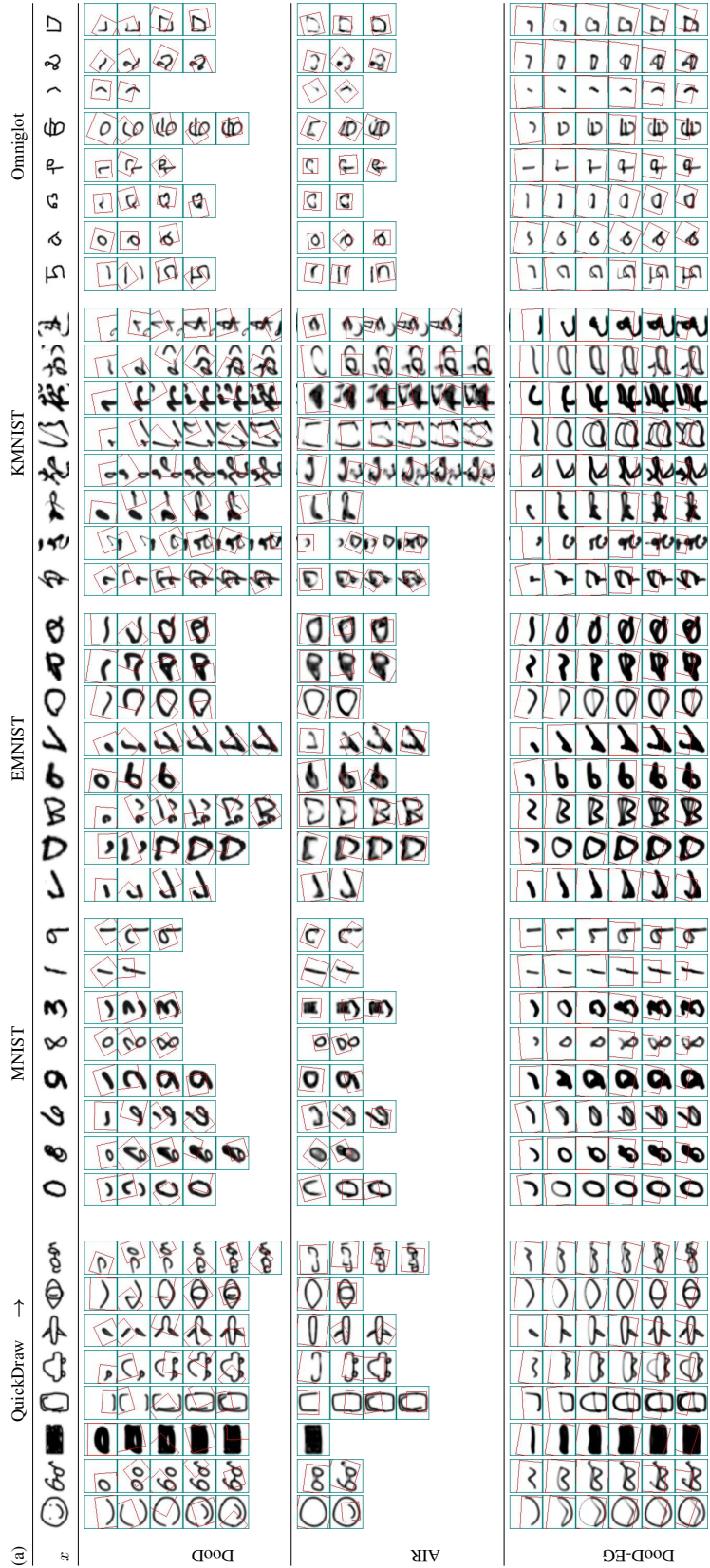
\captionof{figure}{QuickDraw-trained model generalize to other datasets.}
    \label{fig:quickdraw_generalization}
    \end{minipage}
  \end{sideways}
\end{center}

\newpage
\subsubsection{Ablation marginal likelihood evaluation}
\begin{figure}[!h]
  \centering
  \includegraphics[width=0.95\textwidth]{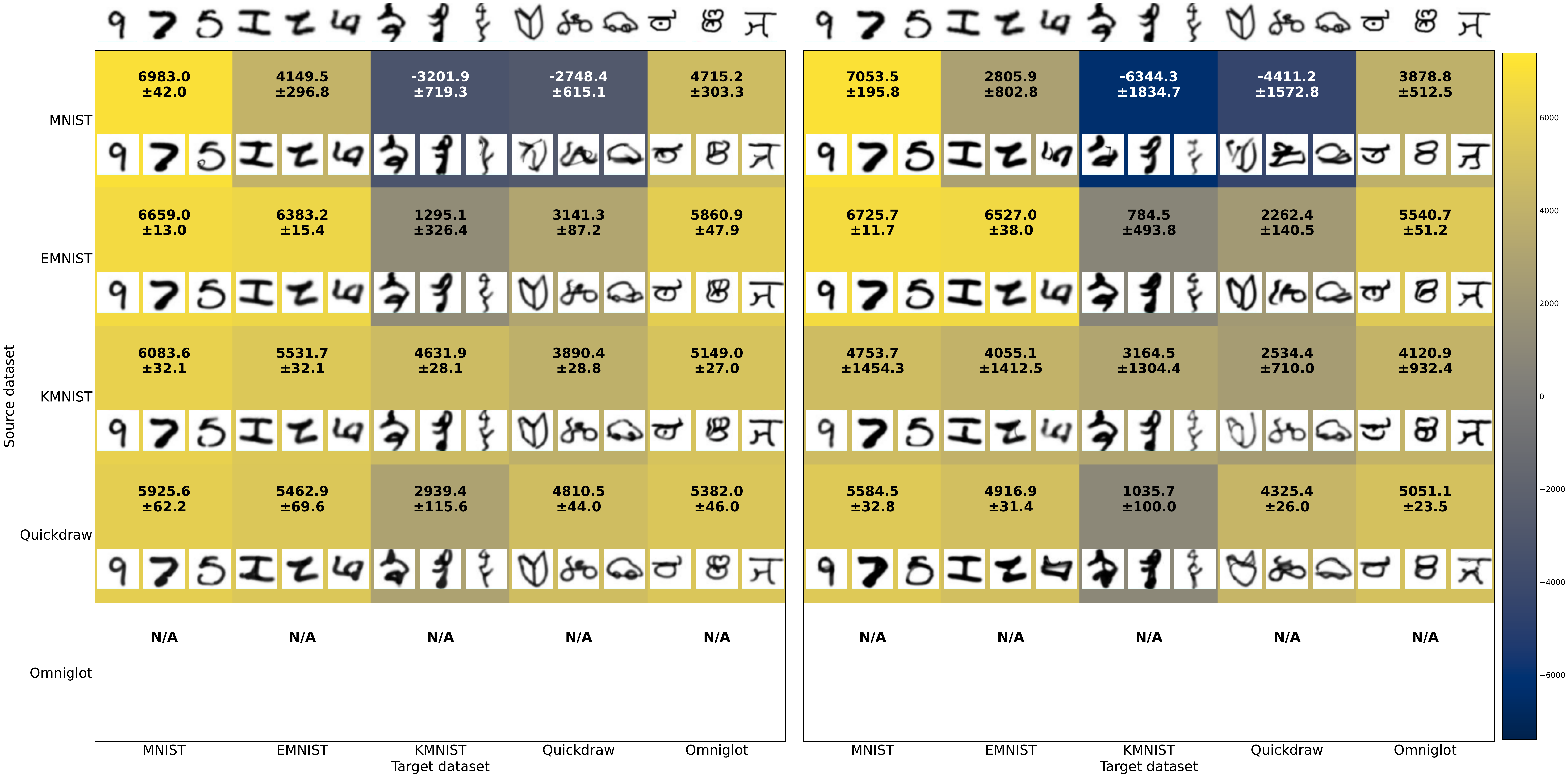}
  \caption{DooD-SP (DooD without sequential prior) and DooD-EG (DooD without execution guidance) cross-dataset log-marginal-likelihood evaluation.}
  \label{fig:cross_ds_mll_eval_dood-eg}
\end{figure}

\newpage
\subsection{More across-task generalization results}

\subsubsection{Unconditional generation}
Additional unconditional samples from DooD are shown in \cref{fig:uncon_gen_app}. In generating these samples, we also make use of the common low-temperature sampling technique \citep{ha2017neural}.

\begin{figure}[!h]
  \def\gencropimg#1{\includegraphics[width=22ex, trim={{1.85\textwidth} 0 0 0},clip]{#1}}
  \def\gencropimgg#1{\includegraphics[width=22ex, trim={{1.0\textwidth} 0 {.85\textwidth} 0},clip]{#1}}
  \def\gencropimggg#1{\includegraphics[width=22ex, trim={0 0 {1.85\textwidth} 0},clip]{#1}}
  \centering\scriptsize
  \scalebox{0.9}{%
  \begin{tabular}{@{}*{5}{c@{\hspace*{10pt}}}@{\hspace*{-10pt}}}
    MNIST
    & EMNIST
    & KMNIST
    & QuickDraw
    & Omniglot\\
    \midrule
    \gencropimg{uncon_generation/uncon_gen_mn.pdf}
    & \gencropimg{uncon_generation/uncon_gen_em.pdf}
    & \gencropimg{uncon_generation/uncon_gen_km.pdf}
    & \gencropimg{uncon_generation/uncon_gen_qd.pdf}
    & \gencropimg{uncon_generation/uncon_gen_om.pdf}\\
    \gencropimgg{uncon_generation/uncon_gen_mn.pdf}
    & \gencropimgg{uncon_generation/uncon_gen_em.pdf}
    & \gencropimgg{uncon_generation/uncon_gen_km.pdf}
    & \gencropimgg{uncon_generation/uncon_gen_qd.pdf}
    & \gencropimgg{uncon_generation/uncon_gen_om.pdf}\\
    \gencropimg{supp_uncon_gen/uncon_gen_mn0.pdf}
    & \gencropimg{supp_uncon_gen/uncon_gen_em0.pdf}
    & \gencropimg{supp_uncon_gen/uncon_gen_km0.pdf}
    & \gencropimg{supp_uncon_gen/uncon_gen_qd0.pdf}
    & \gencropimg{supp_uncon_gen/uncon_gen_om1.pdf}\\
    \gencropimgg{supp_uncon_gen/uncon_gen_mn0.pdf}
    & \gencropimgg{supp_uncon_gen/uncon_gen_em0.pdf}
    & \gencropimgg{supp_uncon_gen/uncon_gen_km0.pdf}
    & \gencropimgg{supp_uncon_gen/uncon_gen_qd0.pdf}
    & \gencropimgg{supp_uncon_gen/uncon_gen_om1.pdf}\\
    \gencropimggg{supp_uncon_gen/uncon_gen_mn0.pdf}
    & \gencropimggg{supp_uncon_gen/uncon_gen_em0.pdf}
    & \gencropimggg{supp_uncon_gen/uncon_gen_km0.pdf}
    & \gencropimggg{supp_uncon_gen/uncon_gen_qd0.pdf}
    & \gencropimggg{supp_uncon_gen/uncon_gen_om1.pdf}\\
    \gencropimg{supp_uncon_gen/uncon_gen_mn1.pdf}
    & \gencropimg{supp_uncon_gen/uncon_gen_em2.pdf}
    & \gencropimg{supp_uncon_gen/uncon_gen_km1.pdf}
    & \gencropimg{supp_uncon_gen/uncon_gen_qd1.pdf}
    & \gencropimg{supp_uncon_gen/uncon_gen_om2.pdf}\\
    \gencropimgg{supp_uncon_gen/uncon_gen_mn1.pdf}
    & \gencropimgg{supp_uncon_gen/uncon_gen_em2.pdf}
    & \gencropimgg{supp_uncon_gen/uncon_gen_km1.pdf}
    & \gencropimgg{supp_uncon_gen/uncon_gen_qd1.pdf}
    & \gencropimgg{supp_uncon_gen/uncon_gen_om2.pdf}\\
    \gencropimgg{supp_uncon_gen/uncon_gen_mn1.pdf}
    & \gencropimgg{supp_uncon_gen/uncon_gen_em2.pdf}
    & \gencropimgg{supp_uncon_gen/uncon_gen_km1.pdf}
    & \gencropimgg{supp_uncon_gen/uncon_gen_qd1.pdf}
    & \gencropimgg{supp_uncon_gen/uncon_gen_om2.pdf}\\
    \gencropimg{supp_uncon_gen/uncon_gen_mn2.pdf}
    & \gencropimg{supp_uncon_gen/uncon_gen_em4.pdf}
    & \gencropimg{supp_uncon_gen/uncon_gen_km2.pdf}
    & \gencropimg{supp_uncon_gen/uncon_gen_qd2.pdf}
    & \gencropimg{supp_uncon_gen/uncon_gen_om3.pdf}\\
    \gencropimgg{supp_uncon_gen/uncon_gen_mn2.pdf}
    & \gencropimgg{supp_uncon_gen/uncon_gen_em4.pdf}
    & \gencropimgg{supp_uncon_gen/uncon_gen_km2.pdf}
    & \gencropimgg{supp_uncon_gen/uncon_gen_qd2.pdf}
    & \gencropimgg{supp_uncon_gen/uncon_gen_om3.pdf}\\
  \end{tabular}}
  \caption{Additional Unconditional generation results from DooD.}
  \label{fig:uncon_gen_app}
\end{figure}

\newpage
\subsubsection{Character-conditioned generation}
\begin{figure}[!h]
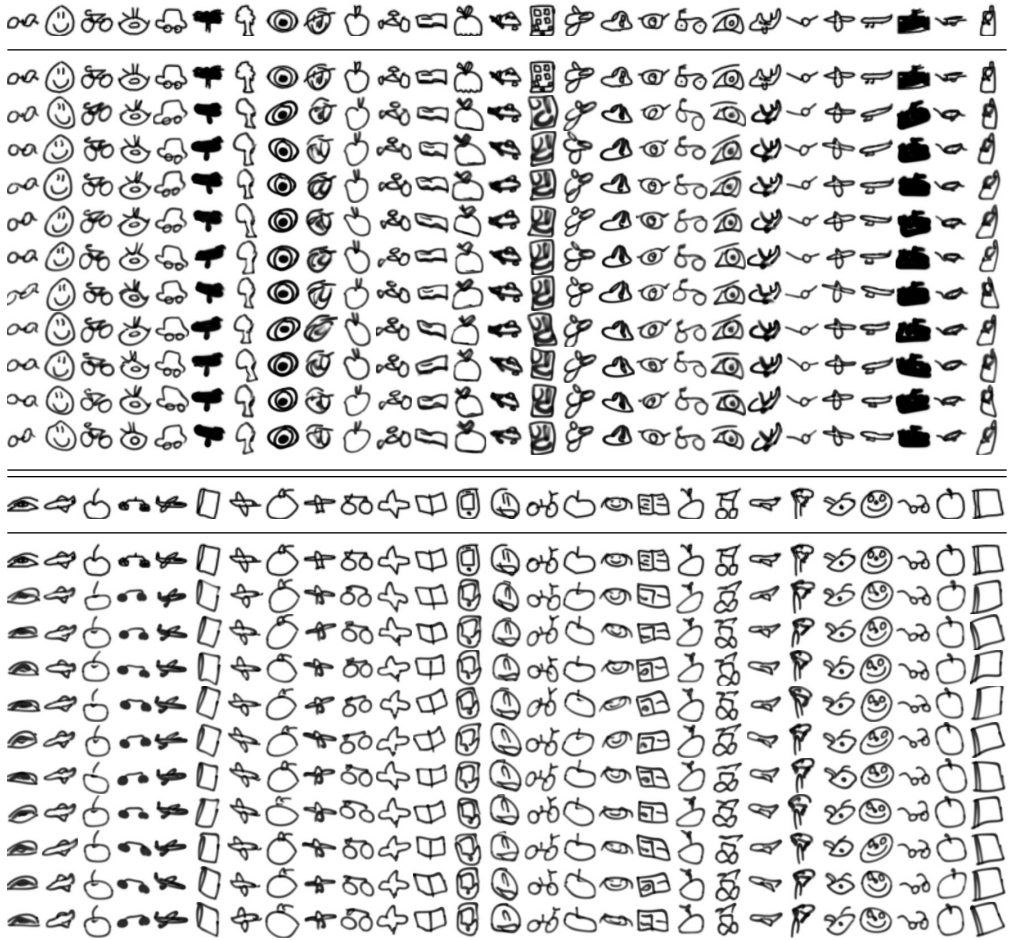

  \def\concropgt#1{\includegraphics[width=84ex, trim={{0.85\textwidth} {1\textheight} 0 0},clip]{#1}}
  \def\concropimg#1{\includegraphics[width=84ex, trim={{0.85\textwidth} 0 0 0},clip]{#1}}
  \centering
  \begin{tabular}{@{}*{1}{c@{\hspace{12pt}}}@{\hspace{-12pt}}}
    \concropgt{char_con_gen/char_con_gen_dood-qd.pdf}\\
    \midrule
    \concropimg{char_con_gen/char_con_gen_dood-qd.pdf}\\
    \midrule
    \midrule
    \concropgt{char_con_gen/char_con_gen_dood-qd2.pdf}\\
    \midrule
    \concropimg{char_con_gen/char_con_gen_dood-qd2.pdf}\\
  \end{tabular}
  \caption{
  Additional Character-conditioned generation results from QuickDraw-trained DooD.
  }
  \label{fig:char_con_gen_app_qd}
\end{figure}

Additional character-conditioned samples from QuickDraw- and Omniglot-trained DooD are shown in \cref{fig:char_con_gen_app_qd} and \cref{fig:char_con_gen_app_om}.

\newpage
\begin{figure}[!h]
  \def\concropgt#1{\includegraphics[width=84ex, trim={{0.85\textwidth} {1\textheight} 0 0},clip]{#1}}
  \def\concropimg#1{\includegraphics[width=84ex, trim={{0.85\textwidth} 0 0 0},clip]{#1}}
  \centering
  \begin{tabular}{@{}*{1}{c@{\hspace{12pt}}}@{\hspace{-12pt}}}
    \concropgt{char_con_gen/char_con_gen_dood-om.pdf}\\
    \midrule
    \concropimg{char_con_gen/char_con_gen_dood-om.pdf}\\
    \midrule
    \midrule
    \concropgt{char_con_gen/char_con_gen_dood-om2.pdf}\\
    \midrule
    \concropimg{char_con_gen/char_con_gen_dood-om2.pdf}\\
  \end{tabular}
  \caption{
  Additional Character-conditioned generation results from Omniglot-trained DooD.
  }
  \label{fig:char_con_gen_app_om}
\end{figure}

\newpage
\subsubsection{One-shot classification}
\label{app:one_shot_classification_details}

DooD performs one-shot classification as follows. When given a support image $\xc$ from each class $c=[1,C]$, and a query image $\xT$, it classifies which class $c$ it belongs to by computing the Bayesian score $p(\xT|\xc)$ for each $c$ and predicting the $c$ with the highest score. The score is computed by:

\begin{align}
    p(\xT|\xc) 
    &= \int p(\xT, \zT, \psic | \xc) d(\zT, \psic)\\
    &\approx \int p(\xT|\zT) p(\zT|\psic) p(\psic|\xc) d(\zT, \psic)\\
    &\approx \sum_{k=1}^K \pi_k\max_{\zT} p(\xT|\zT) p(\zT|\psic_k)\\ &~\text{where}~\psic_k \sim q(\psic|\xc), \pi_k\propto\tilde{\pi}_k=p(\psic_k, \xc)~\text{and}~\sum_{k=1}^K \pi_k = 1
\end{align}
where $p(\zT|\psic)$ is the plug-and-play token model (as introduced in of \cref{app:token_model}) taking the potential affine transformation, motor noise into consideration. And the $\max_{\zT}$ is obtained through gradient-based optimization, as in \citep{lake2015human, feinman2020learning}.

\section{Limitations}
\label{app:limitation}
For MNIST-trained DooD in particular, we observe that despite outperforming all baselines in generalization, as evident by the significantly superior mll (\cref{fig:cross_ds_mll_eval}), it has a hard time faithfully reconstructing in particular the more complex samples.
We attribute this to 2 components of our model that will be investigated in future work, fixing either should significantly improve upon the current generalization performance.

Primarily, we can attribute this to the $\zwhat$-component not generalizing strongly.
When trained on MNIST, the model rarely sees multiple strokes appearing inside a glimpse.
However, this is common in complex dataset such as QuickDraw.
This creates a major train/test discrepancy for the $\zwhat$-component, causing the model's malfunction---e.g., trying to cover 2 isolated strokes (a ``11'') with 1 stroke in the middle (a horizontal bar). 
Despite multiple strokes appearing in glimpses of models trained on other the datasets. This malfunctioning of $\zwhat$-component does not happen as much because the model has more robustly learned through its source dataset that it should focus on the center of the given glimpses during inference.
A more robust $\zwhat$-component design should in principle address this issues.

Fundamentally, however, this can be seen to be caused due to constraints in $\zwhere$.
By design, $\zwhere$ should perfectly segment out each individual stroke and place it into a canonical reference frame, before passing it to the $\zwhat$-component---multiple strokes appearing in a single glimpse should not have happened in the first place.
Perhaps more flexible STNs\citep{jaderberg2015spatial} could do this (with shear, skew, etc).
We expect a combination of the bounding-box approach (as in DooD) and a masking approach (e.g., \citep{burgess2019monet}) might work well, where the masking could help the model ignore irrelevant parts of glimpses before fitting splines to the relevant parts.
\end{document}